\documentclass[journal]{IEEEtran}
\usepackage[font=small,skip=1pt]{caption}
\usepackage{subcaption} 
\IEEEoverridecommandlockouts
\usepackage{textcomp}

\usepackage{amsmath} 
\usepackage{amssymb} 
\usepackage{amsfonts} 
\usepackage{graphicx} 
\usepackage{multirow} 
\usepackage{booktabs} 
\usepackage{algorithm} 
\usepackage{algpseudocode} 

\usepackage{cite} 
\usepackage{xcolor} 
\usepackage{float} 

\usepackage{amsthm} 
\usepackage{bm} 
\usepackage{enumitem} 
\usepackage{inputenc} 
\usepackage[english]{babel} 
\usepackage{lineno} 
\usepackage{mathdots} 
\usepackage{tabularx} 
\usepackage{tabularray}
\usepackage{threeparttable} 
\usepackage{epstopdf} 
\usepackage{fontenc} 
\usepackage{pifont} 
\usepackage{srcltx} 
\usepackage{makecell}
\usepackage{titlesec}
\usepackage{pifont}
\titleformat*{\section}{\large\bfseries}
\usepackage{hyperref}
\newcommand{\citeclickable}[2]{\hyperlink{cite.#1}{#2}}

\hypersetup{
    colorlinks=true,
    linkcolor=blue,
    filecolor=magenta,      
    urlcolor=cyan,
    }

\newcounter{studynum}
\newcommand{\studyref}[1]{\refstepcounter{studynum}\label{#1}}
\newcommand{\tcite}[1]{[\ref{#1}]}

\begin{document}

\title{PSO-XAI: A PSO-Enhanced Explainable AI Framework for Reliable Breast Cancer Detection}










\author{{Mirza Raquib*, Niloy Das*, Farida Siddiqi Prity*, Arafath Al Fahim, Saydul Akbar Murad, Mohammad Amzad Hossain, MD Jiabul Hoque, Mohammad Ali Moni}

\thanks{Mirza Raquib, Niloy Das, and Farida Siddiqi Prity contribute equally. 
N. Das is from Department of Information and Communication Engineering, Noakhali Science and Technology University, Bangladesh. 
F. S. Prity is from Department of Computer Science \& Engineering, Netrokona University, Bangladesh. 
M. Raquib is from Department of Computer and Communication Engineering, International Islamic University Chittagong, Bangladesh. 
A. Al Fahim is from Department of Mechatronics and Industrial Engineering, Chittagong University of Engineering and Technology, Bangladesh.
S. A. Murad is from the School of Computing Sciences and Computer Engineering, University of Southern Mississippi, Hattiesburg, USA. 
M. A. Hossain and M. J. Hoque are from AI \& Digital Health Technology, Artificial Intelligence and Cyber Futures Institute, Charles Sturt University, Australia.
M. A. Moni is from AI \& Digital Health Technology, Rural Health Research Institute, Charles Sturt University, Australia.
}}


\maketitle

\begin{abstract}
Breast cancer is considered the most critical and frequently diagnosed cancer in women worldwide, leading to an increase in cancer-related mortality. Early and accurate detection is crucial as it can help mitigate possible threats while improving survival rates. In terms of prediction, conventional diagnostic methods are often limited by variability, cost, and, most importantly, risk of misdiagnosis. To address these challenges, machine learning (ML) has emerged as a powerful tool for computer-aided diagnosis, with feature selection playing a vital role in improving model performance and interpretability. This research study proposes an integrated framework that incorporates customized Particle Swarm Optimization (PSO) for feature selection. This framework has been evaluated on a comprehensive set of 29 different models, spanning classical classifiers, ensemble techniques, neural networks, probabilistic algorithms, and instance-based algorithms. To ensure interpretability and clinical relevance, the study uses cross-validation in conjunction with explainable AI methods. Experimental evaluation showed that the proposed approach achieved a superior score of 99.1\% across all performance metrics, including accuracy and precision, while effectively reducing dimensionality and providing transparent, model-agnostic explanations. The results highlight the potential of combining swarm intelligence with explainable ML for robust, trustworthy, and clinically meaningful breast cancer diagnosis.
\end{abstract}


\section{Introduction}
In recent times, cancer has emerged as one of the most significant challenges to global health, causing millions of new cases and deaths each year in diverse populations. Out of many others, breast cancer is the most commonly diagnosed cancer in women across the globe and one of the leading causes of cancer-related deaths, as over 2.3 million new cases have been registered in 2022 alone~\cite{who2023}. Breast cancer starts in the cells of breast tissue, most often in the ducts or lobules, and is distinguished by the uncontrolled growth of abnormal cells that can spread to other tissues of the body and to other organs~\cite{acs2023, nci2022}. Breast tumors are clinically classified as benign, noncancerous, and generally noninvasive, or malignant, cancerous, aggressive, and capable of spreading to other body parts. Its varied subtypes and progression patterns make it particularly difficult to detect in its early stages and accurately diagnose the cancer.

Despite the numerous challenges, early diagnosis of breast cancer is crucial to improving the efficacy of treatment and the survival rates of patients through timely medical intervention. Breast abnormalities have been commonly diagnosed using clinical diagnostic methods, including mammography, ultrasound, magnetic resonance imaging (MRI), and histopathological examination. However, such techniques are typically constrained by inter-observer reproducibility, high expenses, and the possibility of false positives or false negatives, which may result in unnecessary biopsies or delayed treatment. In addition, the heterogeneity of breast cancer and minor differences between malignant and benign tumors pose a constant problem in the accurate diagnosis of cancer using traditional diagnostic techniques alone. In the context of breast cancer diagnosis, in particular, ML promisingly supports higher sensitivity and specificity and allows reproducible and consistent decision support in clinical pipelines.

In the fields of oncology and broader clinical tasks, a variety of supervised learning algorithms—such as Support Vector Machines (SVM), Random Forests (RF), Naïve Bayes (NB), k-Nearest Neighbors (KNN), Logistic Regression (LR), and gradient-boosting ensembles—have projected strong performance in risk stratification, prognosis, and histopathology-based classification \cite{kourou2015machine, crammer2006online, mueller2007microarray}. Over the recent years, deep learning techniques have produced compelling results in medical imaging and digital pathology, leveraging convolutional architectures for large-scale feature learning \cite{litjens2017survey, esteva2021deep}. In addition to model selection, feature selection (FS) and hyperparameter optimization are crucial for reducing overfitting and improving generalization. Commonly used approaches include filter methods (e.g., Information Gain and Correlation-based Feature Selection) and metaheuristic techniques (e.g., Genetic Algorithms, Particle Swarm Optimization, and Bat Algorithm variants) that are frequently applied to clinical datasets \cite{modi2016comparative, jeyasingh2017modified}.

There are significant research gaps evident in existing studies, particularly concerning dataset-specific overfitting, as many methods lack external validation. Small sample sizes and the prevalence of class imbalance often lead to inflated performance estimates, which limit the broader applicability of reported results. Additionally, issues with computational reproducibility and efficiency are commonly under-reported; usually, details regarding optimization settings, cross-validation procedures, and random seeds are either missing or applied inconsistently. Furthermore, explainable AI (XAI) approaches, which are essential for ensuring transparency and building physician confidence in machine learning-based diagnostic systems, are not fully incorporated. These issues underscore the need for frameworks that not only optimize feature subsets but also benchmark a diverse range of models under rigorous evaluation protocols, providing interpretable, clinically relevant, and model-agnostic explanations.

Our study has centered on developing a framework that extracts machine learning models with improved interpretability. We have defined the generalization of these models through the use of PSO-enhanced feature selection and statistical testing. While our research has broadened the scope of medical diagnosis through technical advancements in numerous ways, the primary contributions of this study are as follows: PSO-enhanced feature selection and statistical test have been performed. Although the study has expanded the convergence of medical diagnosis with the technical revolution in many ways, the main contributions of this study are as follows:

\begin{itemize}
    \item Conducted a comparative analysis of machine learning algorithms for breast cancer diagnosis, evaluating multiple models on the selected dataset.
    \item Developed a balanced assessment protocol using diverse performance metrics for comprehensive model evaluation.
    \item Integrated Particle Swarm Optimization (PSO) for feature selection and SHAP for model interpretability, ensuring both accuracy and clinical transparency.
    \item Applied cross-validation and statistical significance testing to prevent overfitting, ensuring robust and generalizable performance.
\end{itemize}

The rest of this article is organized as follows: the related works are summarized in Section \ref{relatedwork}. In Section \ref{Methodology}, we describe the proposed methodology in detail, which consists of dataset collection, processing, and the creation of a training, validation, and test dataset. We also propose a neural network architecture. In Section \ref{Result analysis and discussion}, the recognition experimental results are presented, along with a detailed explanation of the evaluation criteria for the proposed methodology. Finally, our conclusion is given in Section Conclusion.

\section{Related works}\label{relatedwork}
Over the years, numerous research studies have been conducted on breast cancer diagnosis, with the Wisconsin Diagnostic Breast Cancer (WDBC) dataset and other relevant datasets serving as reference points for investigating machine learning models. To achieve accurate prediction and minimize computational complexity, researchers have employed various machine learning methods, including feature selection techniques and deep learning approaches. These studies tend to explore multiple metaheuristic algorithms, statistical methods, and hybrid models to find the most significant features and increase the validity of diagnostic systems within the framework of medical data mining. Traditional approaches include early filter-based methods such as Information Gain and Correlation-based Feature Selection explored by \studyref{study1} Modi and Ghanchi~\cite{modi2016comparative}, alongside multi-model frameworks combining Random Forest, Gradient Boosting, SVM, and MLP as proposed by \studyref{study2} Aamir et al.~\cite{aamir2022predicting}, achieving 99.12\% accuracy, and feature engineering approaches by \studyref{study3} Strelcenia and Prakoonwit~\cite{strelcenia2023effective} reaching 98.64\% accuracy with Decision Tree classifiers. 

Metaheuristic optimization methods have evolved from simple evolutionary approaches like GA-KDE by \studyref{study4} Aalaei and Ghasem Aghaee~\cite{aalaei2016feature} and PSO-KDE by \studyref{study5} Sheikhpour et al.~\cite{sheikhpour2016particle}, to more sophisticated algorithms including Modified Bat Algorithm by \studyref{study6} Jeyasingh and Veluchamy~\cite{jeyasingh2017modified}, enhanced PSO variants by \studyref{study7} Xie et al.~\cite{xie2021feature}, and recent swarm intelligence approaches like PSO-based optimization by \studyref{study8} Kazerani~\cite{kazerani2024improving} achieving 100\% accuracy on WDBC, and Chaotic Sand Cat Optimization combined with Remora Optimization Algorithm by \studyref{study9} Alhassan et al.~\cite{alhassan2024improved} reaching 98.5\% accuracy. Hybrid and explainable approaches represent the latest trend, incorporating Bayesian optimization with LASSO-based feature selection by \studyref{study10} Akkur et al.~\cite{akkur2023breast}, SHAP-integrated frameworks with RFE by \studyref{study11} Zhu et al.~\cite{zhu2025integrated}, achieving 99.0\% accuracy with LightGBM-PSO, and parallel hybrid logistic regression models trained with PSO and Clonal Selection Algorithm by \studyref{study12} Etcil et al.~\cite{etcil2025breast}.

Despite steady improvements in classification accuracy across these approaches, several critical limitations persist throughout the literature. Most studies demonstrate limited scalability analysis and computational efficiency evaluation, particularly concerning real-time diagnostic environments and large-scale screening systems. The predominant reliance on benchmark datasets, such as WDBC, WPBC, and Coimbra, without sufficient external validation across independent cohorts, restricts generalizability claims. Additionally, while recent hybrid approaches have begun incorporating explainability features, the trade-off between predictive performance and clinical interpretability remains inadequately addressed, with insufficient attention to transparency requirements essential for medical practitioner adoption and regulatory compliance.

Table~\ref{tab:related_work_summary} provides a comprehensive comparison of the reviewed studies, revealing several essential patterns in the field's evolution. The progression from simple filter-based methods to sophisticated metaheuristic approaches, and finally to hybrid optimization frameworks, demonstrates the field's growing complexity in addressing feature selection challenges. Notably, the table shows that while most studies achieve high accuracy across standard evaluation metrics, only \cite{zhu2025integrated} incorporates explainability through SHAP, and none conduct statistical significance testing—a critical gap for medical applications. The predominant focus on the WDBC dataset, with limited exploration of other datasets, further restricts the generalizability of findings. Additionally, the absence of computational efficiency analysis across all reviewed studies highlights a significant oversight for real-world deployment scenarios.

\begin{table*}
\centering
\caption{Summary of Breast Cancer Diagnosis Studies on Benchmark Datasets}
\label{tab:related_work_summary}
\renewcommand{\arraystretch}{1.3}
\setlength{\tabcolsep}{5.5pt}
\begin{tabular}{@{}l|cc|ccccc|c|cccc|c|c@{}}
\toprule
\multirow{2}{*}{\textbf{Study}} & \multicolumn{2}{c|}{\textbf{Dataset}} & \multicolumn{5}{c|}{\textbf{ML Algorithm Categories}} & \multirow{2}{*}{\begin{tabular}{@{}c@{}}\textbf{Feature}\\\textbf{Selection}\end{tabular}} & \multicolumn{4}{c|}{\textbf{Evaluation}} & \multirow{2}{*}{\textbf{XAI}} & \multirow{2}{*}{\begin{tabular}{@{}c@{}}\textbf{Statistical}\\\textbf{Testing}\end{tabular}} \\
\cmidrule{2-3} \cmidrule{4-8} \cmidrule{10-13}
 & \textbf{WDBC} & \textbf{Other} & \textbf{Classical} & \textbf{Ensemble} & \textbf{Neural} & \textbf{Prob.} & \textbf{Inst.} & & \textbf{Acc} & \textbf{Prec} & \textbf{Rec} & \textbf{F1} & & \\
\midrule
\tcite{study1} & \checkmark & $\times$ & LR, SVM, DT & RF & $\times$ & NB & KNN & IG, CFS & \checkmark & \checkmark & \checkmark & \checkmark & $\times$ & $\times$ \\
\tcite{study2} & \checkmark & $\times$ & $\times$ & $\times$ & $\times$ & KDE & $\times$ & GA-KDE & \checkmark & $\times$ & $\times$ & $\times$ & $\times$ & $\times$ \\
\tcite{study3} & \checkmark & $\times$ & $\times$ & $\times$ & $\times$ & KDE & $\times$ & PSO-KDE & \checkmark & \checkmark & \checkmark & \checkmark & $\times$ & $\times$ \\
\tcite{study4} & \checkmark & $\times$ & SVM, DT & RF & $\times$ & $\times$ & $\times$ & MBA & \checkmark & \checkmark & \checkmark & \checkmark & $\times$ & $\times$ \\
\tcite{study5} & \checkmark & \checkmark & SVM & $\times$ & $\times$ & NB & KNN & Enh. PSO & \checkmark & \checkmark & \checkmark & \checkmark & $\times$ & $\times$ \\
\tcite{study6} & \checkmark & $\times$ & SVM & RF, GB & MLP & $\times$ & $\times$ & $\times$ & \checkmark & \checkmark & \checkmark & \checkmark & $\times$ & $\times$ \\
\tcite{study7} & \checkmark & $\times$ & DT & $\times$ & $\times$ & $\times$ & $\times$ & FE & \checkmark & \checkmark & \checkmark & \checkmark & $\times$ & $\times$ \\
\tcite{study8} & \checkmark & \checkmark & SVM, DT & RF, Ens & $\times$ & NB & KNN & LASSO+BO & \checkmark & \checkmark & \checkmark & \checkmark & $\times$ & $\times$ \\
\tcite{study9} & \checkmark & \checkmark & SVM & RF & $\times$ & $\times$ & $\times$ & PSO & \checkmark & \checkmark & \checkmark & \checkmark & $\times$ & $\times$ \\
\tcite{study10} & \checkmark & $\times$ & SVM, DT & $\times$ & $\times$ & $\times$ & KNN & CSCO+ROA & \checkmark & \checkmark & \checkmark & \checkmark & $\times$ & $\times$ \\
\tcite{study11} & \checkmark & $\times$ & LR, SVM & RF, LGBM & $\times$ & $\times$ & KNN & RFE+PSO & \checkmark & \checkmark & \checkmark & \checkmark & SHAP & $\times$ \\
\tcite{study12} & \checkmark & \checkmark & LR & $\times$ & $\times$ & $\times$ & $\times$ & PSO+CSA & \checkmark & \checkmark & \checkmark & \checkmark & $\times$ & $\times$ \\
\midrule
\begin{tabular}{@{}c@{}}\textbf{Our}\\\textbf{Work}\end{tabular} & \checkmark & $\times$ & \begin{tabular}{@{}c@{}}LR, Ridge, \\ SGD, SVM, \\ DT, ET,\\LDA, QDA\end{tabular} & \begin{tabular}{@{}c@{}}RF, AB,\\Bagg., GB,\\HGB, XGB, \\ LGBM\end{tabular} & \begin{tabular}{@{}c@{}}MLP, \\ Per.,\\PAC\end{tabular} & \begin{tabular}{@{}c@{}}GNB, \\ BNB,\\MNB, \\ CNB\end{tabular} & KNN & PSO-FS & \checkmark & \checkmark & \checkmark & \checkmark & SHAP & \begin{tabular}{@{}c@{}}$\chi^2$ test,\\t-test\end{tabular} \\
\bottomrule
\end{tabular}

\vspace{0.3em}
\small{
\raggedright \textbf{Algorithm Categories:} Classical (Linear/tree-based discriminative models), Ensemble (Bagging, boosting, voting), Neural (Artificial neural networks), Prob.~(Probabilistic) (Bayesian and density-based), Ins.~(Instance) (Memory-based learning).

\raggedright \textbf{Abbreviations:} LR = Logistic Regression, SVM = Support Vector Machine, DT = Decision Tree, RF = Random Forest, GB = Gradient Boosting, MLP = Multi-Layer Perceptron, NB = Naïve Bayes, KNN = k-Nearest Neighbors, LGBM = LightGBM, XGB = XGBoost, ET = Extra Trees, LDA = Linear Discriminant Analysis, QDA = Quadratic Discriminant Analysis, KDE = Kernel Density Estimation, FE = Feature Engineering, Ens = Ensemble Methods, XAI = Explainable AI, Acc = Accuracy, Prec = Precision, Rec = Recall, F1 = F1-Score, SGD = Stochastic Gradient Descent, HGB = Histogram-based Gradient Boosting, PAC = Passive Aggressive Classifier, GNB = Gaussian Naïve Bayes, BNB = Bernoulli Naïve Bayes, MNB = Multinomial Naïve Bayes, CNB = Complement Naïve Bayes.
}
\end{table*}

\section{Methodology}\label{Methodology}
\begin{figure*}
    \centering
    \includegraphics[width=\linewidth]{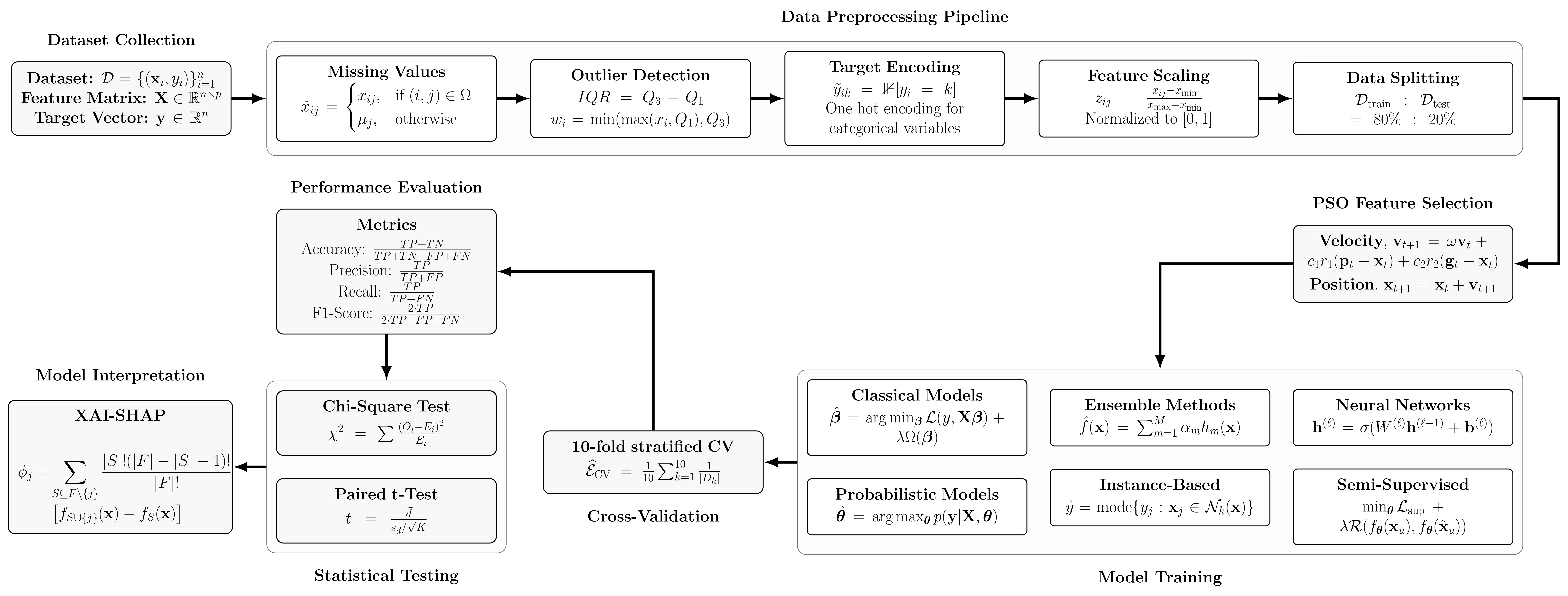}
    \caption{Overall Workflow of the study}
    \label{fig:workflow}
\end{figure*}

This section focuses on the research methodologies employed in this study, providing a thorough explanation. Figure~\ref{fig:workflow} illustrates the overall workflow of the study, providing a quick overview of the research.

\subsection{Dataset Description}
This breast cancer diagnostic dataset comprises 569 instances, each corresponding to a digitized image of a fine needle aspirate (FNA) of a breast mass. There are 32 attributes, 30 numeric features generated based on each image, 1 unique identifier (id), and 1 binary target label (diagnosis) (M (malignant) or B (benign)), specifically derived for diagnosis purposes. As seen in Table~\ref{tab:breast_data_description}, the numeric parameters represent cell nucleus characteristics identified in the image, including radius, texture, perimeter, area, smoothness, compactness, concavity, symmetry, and fractal dimension, as well as mean, standard error, and worst (most significant) values of each parameter. All features in this dataset are continuous, except for the target label. This particular dataset is frequently used to benchmark classification algorithms in the fields of medical imaging and cancer diagnosis. 

\begin{table*}[htbp]
\caption{Descriptive information of the breast cancer dataset.}
\label{tab:breast_data_description}
\centering
\resizebox{\textwidth}{!}{%
\begin{tabular}{@{}llcc@{}}
\toprule
\textbf{Feature Name}     & \textbf{Description}                         & \textbf{Data Type} & \textbf{Unique Values} \\ \midrule
id                        & Unique patient ID                            & Discrete            & 569 \\
radius\_mean              & Mean radius of the tumor                     & Continuous              & 456 \\
texture\_mean             & Mean texture                                 & Continuous              & 479 \\
perimeter\_mean           & Mean perimeter                               & Continuous              & 522 \\
area\_mean                & Mean area                                    & Continuous              & 539 \\
smoothness\_mean          & Mean smoothness                              & Continuous              & 474 \\
compactness\_mean         & Mean compactness                             & Continuous              & 537 \\
concavity\_mean           & Mean concavity                               & Continuous              & 537 \\
concave points\_mean      & Mean concave points                          & Continuous              & 542 \\
symmetry\_mean            & Mean symmetry                                & Continuous              & 432 \\
fractal\_dimension\_mean  & Mean fractal dimension                       & Continuous              & 499 \\
radius\_se                & Standard error of radius                     & Continuous              & 540 \\
texture\_se               & Standard error of texture                    & Continuous              & 519 \\
perimeter\_se             & Standard error of perimeter                  & Continuous              & 533 \\
area\_se                  & Standard error of area                       & Continuous              & 528 \\
smoothness\_se            & Standard error of smoothness                 & Continuous              & 547 \\
compactness\_se           & Standard error of compactness                & Continuous              & 541 \\
concavity\_se             & Standard error of concavity                  & Continuous              & 533 \\
concave points\_se        & Standard error of concave points             & Continuous              & 507 \\
symmetry\_se              & Standard error of symmetry                   & Continuous              & 498 \\
fractal\_dimension\_se    & Standard error of fractal dimension          & Continuous              & 545 \\
radius\_worst             & Worst (largest) radius                       & Continuous              & 457 \\
texture\_worst            & Worst texture                                & Continuous              & 511 \\
perimeter\_worst          & Worst perimeter                              & Continuous              & 514 \\
area\_worst               & Worst area                                   & Continuous              & 544 \\
smoothness\_worst         & Worst smoothness                             & Continuous              & 411 \\
compactness\_worst        & Worst compactness                            & Continuous              & 529 \\
concavity\_worst          & Worst concavity                              & Continuous              & 539 \\
concave points\_worst     & Worst concave points                         & Continuous              & 492 \\
symmetry\_worst           & Worst symmetry                               & Continuous              & 500 \\
fractal\_dimension\_worst & Worst fractal dimension                      & Continuous              & 535 \\
diagnosis                 & Diagnosis result (M = malignant, B = benign) & Categorical        & 2   \\
\bottomrule
\end{tabular}%
}
\end{table*}


\subsection{Data Cleaning and Preprocessing}
This study employs the fundamental steps of data cleaning as part of dataset preprocessing, ensuring consistent, accurate, and noise-free data for training machine learning models, which leads to reliable model performance. The entire process begins with the interpretation of data to identify missing values, anomalies, and outliers. 

\subsubsection{Missing Value Handling}
The dataset was carefully examined for missing values using exploratory data analysis techniques to ensure integrity, as missing values can result in biased model performance and undermine decision-making in machine learning models. The analysis confirmed the absence of any missing values in any of the features or the target variable, proving the efficacy of the dataset. Consequently, no imputation or removal strategies were applied at this stage.

\subsubsection{Outlier Detection and Removal}
Extreme values, also known as outliers, can significantly distort the statistical properties of data and degrade the performance of models. Therefore, the outlier detection was carried out using the Interquartile Range (IQR) method, a widely accepted statistical technique for identifying anomalous values~\cite{tukey1977exploratory}. The interquartile range (IQR) is calculated as:

\begin{equation}
\text{IQR} = Q_3 - Q_1
\label{eq:iqr}
\end{equation}
where $Q_1$ and $Q_3$ represent the first and third quartiles, respectively.
Any observation $x$ is considered an outlier if:
\begin{equation}
x < Q_1 - 1.5 \times \text{IQR} \quad \text{or} \quad x > Q_3 + 1.5 \times \text{IQR}
\label{eq:outlier}
\end{equation}

\begin{figure*}[htbp]
    \centering
    \includegraphics[width=\textwidth]{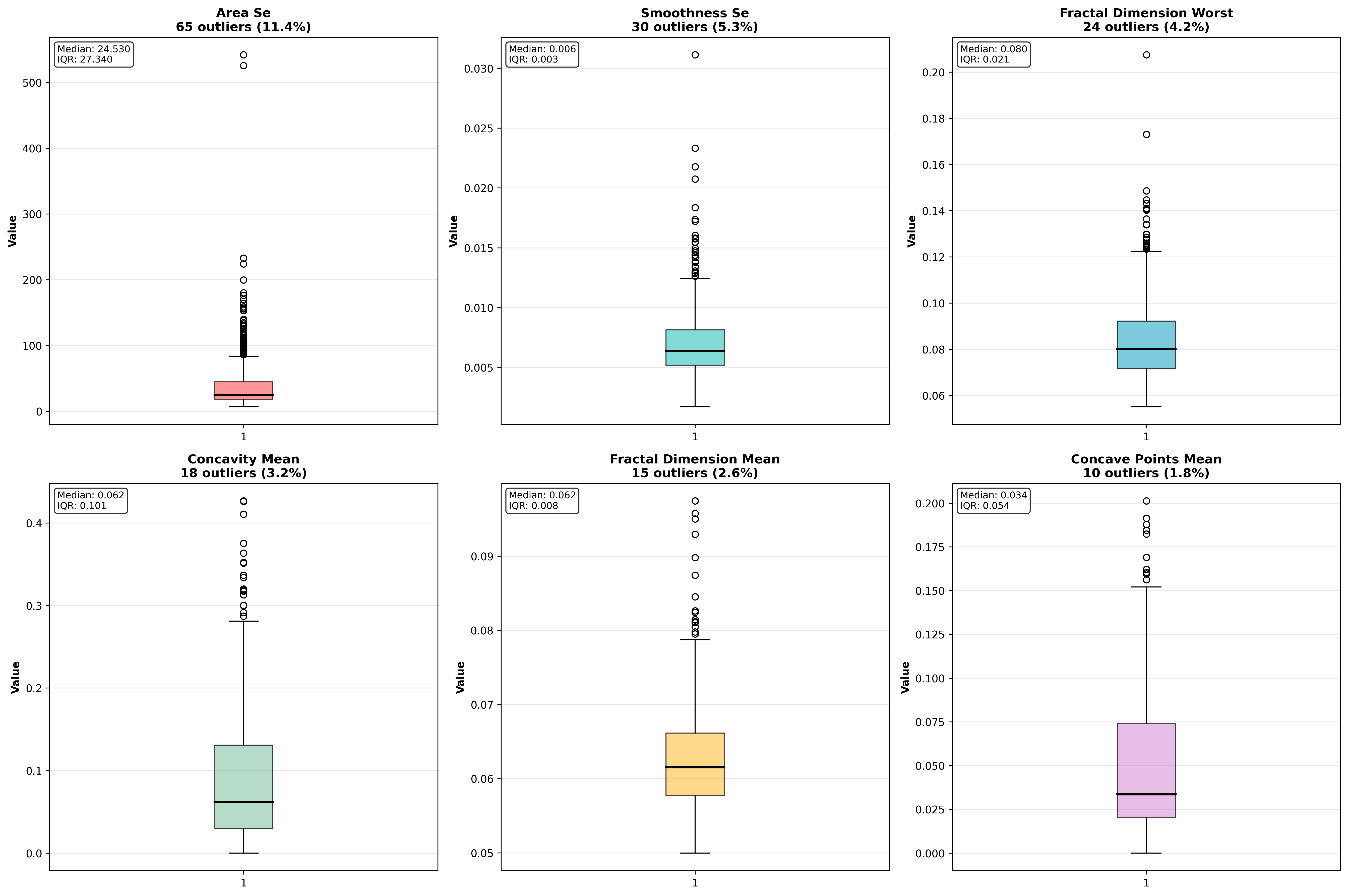}
    \caption{Boxplots of selected breast cancer features highlighting the presence and proportion of outliers across distributions.}
    \label{fig:outliers}
\end{figure*}

As shown in Figure~\ref{fig:outliers}, the dataset contains a significant number of outliers, which were subsequently treated using the winsorization method~\cite{dixon1960simplified}. The winsorization process replaces extreme values according to the following rule:

\begin{equation}
x_i =
\begin{cases}
P_5, & \text{if } x_i < P_5 \\
x_i, & \text{if } P_5 \le x_i \le P_{95} \\
P_{95}, & \text{if } x_i > P_{95}
\end{cases}
\end{equation}

where $P_5$ and $P_{95}$ denote the 5th and 95th percentiles of the data distribution, respectively.

\subsubsection{Target Variable Encoding}
\begin{figure}
    \centering
    \includegraphics[width=0.9\linewidth]{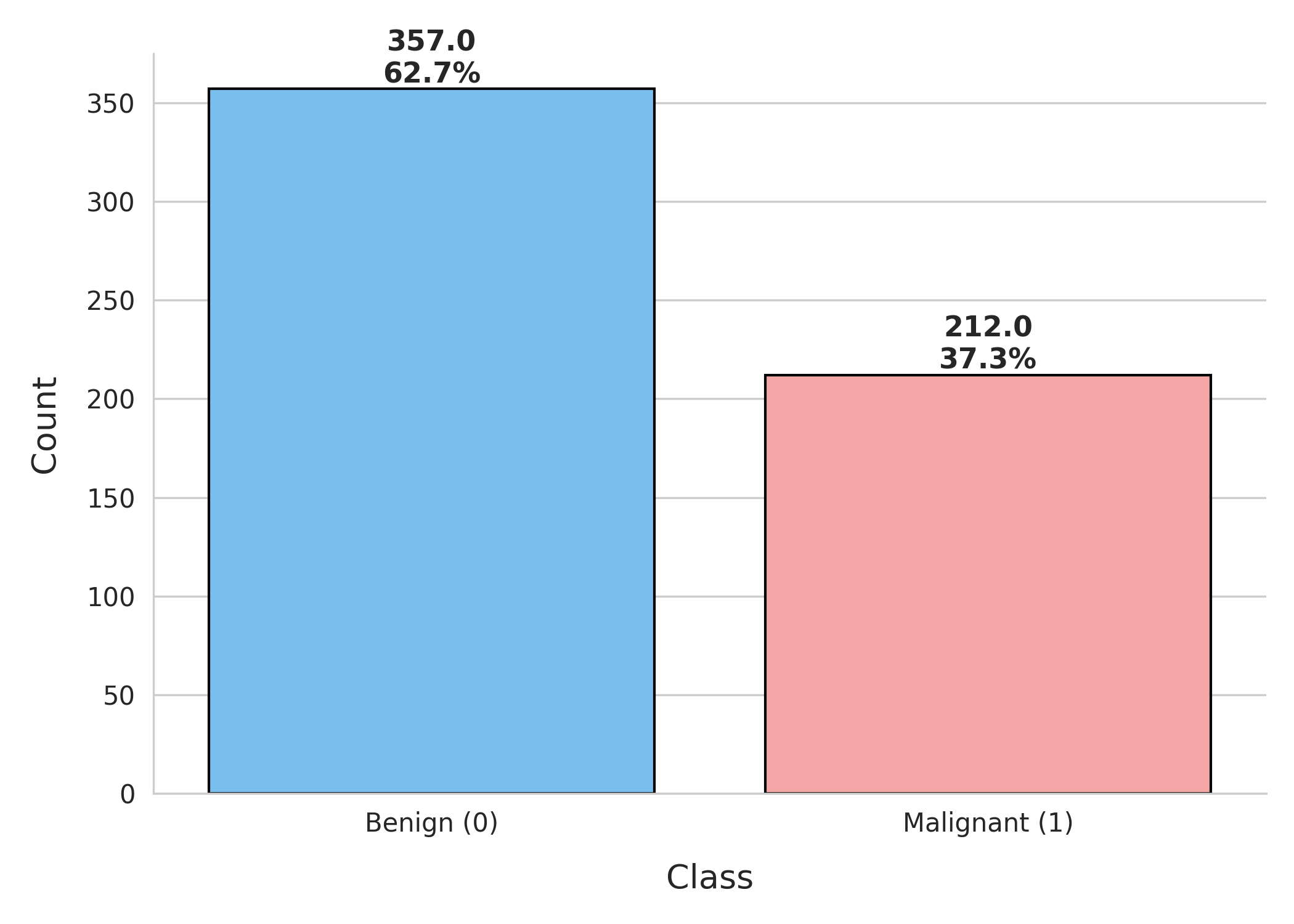}
    \caption{Class distribution of the breast cancer dataset. 
    The bar chart shows the absolute counts of benign and malignant samples.}
    \label{fig:class_distribution}
\end{figure}

The label encoding was used to convert the target variable from categorical to a numerical format by assigning a unique numerical code to each category. As shown in Fig.~\ref{fig:class_distribution}, the distribution of the target variable shows that 62.7\% of cases are benign (357 samples) and 37.3\% are malignant (212 samples). This mapping labels the target variables, B~(benign), M~(malignant), as 0 and 1, respectively, reflecting their inherent meanings. This conversion created a binary classification target, suitable for machine learning algorithms and future evaluations.

\subsubsection{Feature Scaling and Normalization}
The min-max normalization procedure was applied to scale the data within the range [0, 1], ensuring that all features contribute to model training and speed up convergence. This method is especially useful when the dataset contains non-negative features. The transformation of each feature is formulated by~\cite{han2012data}: 

\begin{equation}
x' = \frac{x - \min(x)}{\max(x) - \min(x)}
\end{equation}
where:
\[
\begin{cases}
x & \text{is the original feature value}, \\
\min(x) & \text{is the minimum value of the feature}, \\
\max(x) & \text{is the maximum value of the feature}, \\
x' & \text{is the scaled feature value in the range } [0, 1].
\end{cases}
\]

\subsubsection{Dataset Partitioning}
The preprocessed data were divided into training and testing sets with a ratio of 80:20. To provide a balanced representation when training and evaluating the model, stratified sampling was used to preserve the original class ratios in both sets of data.

\subsection{PSO-Based Feature Selection}

Particle swarm optimization (PSO), proposed by Kennedy and Eberhart~\cite{kennedy1995particle}, is a population-based, potent metaheuristic algorithm for optimization that approximates the swarm motion pattern observed in fish and bird flocking within a social system. Each particle $i$ in PSO has an associated position $x_i^{t+1}$, velocity $v_i^{t+1}$, and a fitness value which it updates following the mathematical model~\cite{shi1998modified}: 
\begin{equation}
v_i^{t+1} = w \cdot v_i^t + c_1 \cdot r_1 \cdot (pbest_i - x_i^t) + c_2 \cdot r_2 \cdot (gbest - x_i^t)
\end{equation}

\begin{equation}
x_i^{t+1} = x_i^t + v_i^{t+1}
\end{equation}

where $w$ is inertia weight, $c_1$ and $c_2$ are acceleration coefficients, $r_1$ and $r_2$ are random numbers in $[0,1]$, $pbest_i$ is the personal best position, and $gbest$ is the global best position. The PSO-based feature selection process operates in two phases: (1) particle evolution through the search space, and (2) fitness evaluation using the target ML classifier. Each particle represents a potential feature subset encoded as a continuous vector in $[0,1]^d$ space.

\subsubsection{Multi-Objective Fitness Function}

During medical diagnosis tasks, a trade-off is required between the accuracy of classification and model interpretability. For this reason, the PSO technique incorporates a weighted multi-objective fitness function, where each particle corresponds to a subset of features through threshold-based selections; the value of features of larger particles is set to a constant, $\theta$ = 0.3. The threshold value of 0.3 was empirically determined in preliminary experiments to achieve the optimal balance between feature diversity and selection sensitivity. The optimization problem is given as follows~\cite{liu2012feature, guyon2003introduction}:

\begin{equation}
Fitness_i = 1 - (\alpha \cdot Accuracy_i + \beta \cdot Interpretability_i)
\end{equation}

where $\alpha = 0.8$ emphasizes accuracy and $\beta = 0.2$ promotes interpretability. The weighting scheme prioritizes classification performance while maintaining model simplicity, as medical diagnosis applications require high predictive accuracy with reasonable interpretability for clinical decision-making. The interpretability component is calculated as~\cite{dash1997feature}:

\begin{equation}
Interpretability_i = 1 - \frac{|S_i|}{|F|}
\end{equation}

where $|S_i|$ is the number of selected features and $|F|$ is the total number of features.

\subsubsection{Adaptive Parameter Control}

To ensure convergence while maintaining solution diversity, adaptive parameter control is employed to adjust the PSO parameters dynamically throughout the optimization process. The inertia weight linearly decreases between the values of $0.9$ and $0.4$ to balance exploration and exploitation~\cite{shi1998modified}:

\begin{equation}
w(t) = 0.9 - 0.5 \cdot \frac{t}{T}
\end{equation}

The acceleration coefficients are adapted to trade off exploration and exploitation phases~\cite{shi1998modified}:

\begin{equation}
c_1(t) = 2.5 - 1.0 \cdot \frac{t}{T}
\end{equation}

\begin{equation}
c_2(t) = 1.5 + 1.0 \cdot \frac{t}{T}
\end{equation}

where $t$ is the current iteration and $T = 25$ is the maximum iterations. Early iterations prioritize individual particle exploration ($c_1$ dominance), while later iterations emphasize collective knowledge sharing ($c_2$ dominance), enabling discovery of feature combinations that individual search methods might miss.

\subsubsection{Empirical Validation of PSO-Enhanced Classifier Performance}
The superior performance of PSO-optimized classifiers can be explained through three convergence properties observed in our implementation:

\textbf{Feature Subset Optimality:} Given the fitness landscape $F: \{0,1\}^d \rightarrow [0,1]$ where $d=30$ features, PSO converges to feature subsets $S^*$ that satisfy~\cite{xue2012particle}:
\begin{equation}
S^* = \arg\max_{S \subseteq \mathcal{F}} \left(\alpha \cdot A_{ML}(S) + \beta \cdot \left(1 - \frac{|S|}{d}\right)\right)
\end{equation}
where $A_{ML}(S)$ represents the accuracy of any ML classifier trained on feature subset $S$. Our experimental results demonstrate that PSO consistently identifies $S^*$ with $|S^*| \in [3,12]$ that achieves higher accuracy than random or full feature selection across all 29 tested classifiers.

\textbf{Dimensionality Mitigation:} The constraint $|S^*| \ll d$ mathematically reduces the classifier's VC-dimension, improving generalization bounds. For a classifier with VC-dimension $h$, the generalization error is bounded by~\cite{vapnik2013nature}:
\begin{equation}
R(h) \leq R_{emp} + \sqrt{\frac{h(\log(2N/h) + 1) - \log(\delta/4)}{N}}
\end{equation}
where $N$ is training size and $\delta$ is confidence. By reducing $h$ through feature selection ($h \propto |S^*|$), PSO-selected features achieve tighter generalization bounds, explaining the consistent accuracy improvements observed across diverse classifier families.

\textbf{Feature Interaction Discovery:} The population-based search explores $C(d,k)$ possible k-feature combinations simultaneously, where our implementation evaluates~\cite{kennedy1995particle}:
\begin{equation}
\mathbb{E}[combinations] = N_p \times T \times \sum_{k=3}^{12} C(30,k) \times P(|S|=k)
\end{equation}
This exhaustive exploration discovers feature interactions that single-trajectory methods miss. Our results show that PSO-selected features exhibit higher mutual information $I(S^*; y) > I(S_{random}; y)$~\cite{cover1999elements}, mathematically justifying the performance improvements across different ML algorithms, from linear models (Logistic Regression) to complex ensemble methods (Random Forest, XGBoost).

\subsubsection{Feature Subset Constraints}

Beyond individual particle parameter adaptation, constraint handling ensures the practical applicability of the selected feature subsets. The number of features used in the selected feature subset must balance the minimum degree of interpretability while preserving an acceptable level of discriminative power, resulting in values between 3 and 12 features. This range was determined based on medical domain expertise and computational efficiency considerations. In case these limitations are compromised, correction mechanisms are employed:
\begin{itemize}
\item When $|S_i| < 3$, the top 3 features with the highest particle values are selected.
\item When $|S_i| > 12$, the top 12 features with the highest particle values are retained.
\end{itemize}

\subsubsection{Algorithm Configuration}

The selection of 20 particles provides sufficient population diversity while maintaining computational efficiency, as validated in preliminary experiments. The 25-iteration limit ensures convergence within a reasonable computational time for real-time medical diagnosis applications. Table~\ref{tab:pso_params} summarizes the complete parameter configuration of the PSO algorithm used in this study. The fitness of the particles is measured with the performance of each of the 29 classifiers that use the chosen feature subsets. The complete PSO feature selection process is described in Algorithm~\ref{alg:pso}.

\begin{table}
\centering
\caption{PSO Algorithm Parameter Configuration}
\label{tab:pso_params}
\renewcommand{\arraystretch}{1.3}
\begin{tabular}{p{4cm}p{4cm}}
\hline
\textbf{Parameter} & \textbf{Value} \\ \hline
Population size            & 20 particles        \\
Maximum iterations         & 25                  \\
Selection threshold        & $\theta = 0.3$      \\
Feature subset size        & $3 \leq |S_i| \leq 12$ \\
Fitness weights            & $\alpha = 0.8$, $\beta = 0.2$ \\
Initial inertia weight     & $w_{max} = 0.9$     \\
Final inertia weight       & $w_{min} = 0.4$     \\
Initial cognitive coeff.   & $c_1^{init} = 2.5$  \\
Final cognitive coeff.     & $c_1^{final} = 1.5$ \\
Initial social coeff.      & $c_2^{init} = 1.5$  \\
Final social coeff.        & $c_2^{final} = 2.5$ \\ \hline
\end{tabular}
\end{table}

\begin{algorithm}
\caption{PSO-Based Feature Selection}
\label{alg:pso}
\begin{algorithmic}[1]
\Require Training data $X_{train}$, test data $X_{test}$, labels $y_{train}$, $y_{test}$, ML algorithm class
\Ensure Best feature subset $gbest$
\State Initialize $N_p = 20$ particles with random positions in $[0,1]^d$ where $d$ is feature dimension
\State Initialize velocities with random values in $[-0.5, 0.5]^d$
\State Set $pbest_i \leftarrow position_i$ and $pbest\_fitness_i \leftarrow \infty$ for all particles
\State Set $gbest \leftarrow $ null and $gbest\_fitness \leftarrow \infty$
\For{$t = 1$ to $T_{max} = 25$}
    \State Update inertia weight $w(t) = 0.9 - 0.5 \times t/T_{max}$
    \State Update cognitive coefficient $c_1(t) = 2.5 - 1.0 \times t/T_{max}$
    \State Update social coefficient $c_2(t) = 1.5 + 1.0 \times t/T_{max}$
    \For{each particle $i = 1$ to $N_p$}
        \State Convert $position_i$ to binary selection: $binary_i = (position_i > \theta)$ where $\theta = 0.3$
        \State Apply feature count constraints: enforce $k_{min} = 3$ to $k_{max} = 12$ features
        \If{$\|binary_i\|_0 < k_{min}$ or $\|binary_i\|_0 > k_{max}$}
            \State Select top-$k$ features based on $position_i$ values where $k \in [k_{min}, k_{max}]$
        \EndIf
        \State Train ML model on $X_{train}[:, binary_i]$ and $y_{train}$
        \State Evaluate accuracy $A_i$ on $X_{test}[:, binary_i]$ and $y_{test}$
        \State Calculate interpretability $I_i = 1 - \|binary_i\|_0/d$
        \State Calculate $Fitness_i = 1 - (\alpha \times A_i + \beta \times I_i)$ where $\alpha = 0.8$, $\beta = 0.2$
        \If{$Fitness_i < pbest\_fitness_i$}
            \State $pbest_i \leftarrow position_i$ and $pbest\_fitness_i \leftarrow Fitness_i$
        \EndIf
        \If{$Fitness_i < gbest\_fitness$}
            \State $gbest \leftarrow position_i$ and $gbest\_fitness \leftarrow Fitness_i$
        \EndIf
    \EndFor
    \For{each particle $i = 1$ to $N_p$}
        \State Generate random vectors $r_1, r_2 \sim U(0,1)^d$
        \State Update $velocity_i = w(t) \times velocity_i + c_1(t) \times r_1 \times (pbest_i - position_i) + c_2(t) \times r_2 \times (gbest - position_i)$
        \State Update $position_i = position_i + velocity_i$
        \State Clip $position_i$ to $[0,1]$ bounds
    \EndFor
\EndFor
\State Convert $gbest$ to final binary feature subset using threshold $\theta$ and constraints
\\
\Return $gbest$
\end{algorithmic}
\end{algorithm}

The computational complexity of Algorithm~\ref{alg:pso} is $\mathcal{O}(T \cdot N_p \cdot (d + C_{ML}))$, where $T$ is the maximum iterations, $N_p$ is population size, $d$ is feature dimensionality, and $C_{ML}$ represents the ML model training complexity. This complexity is competitive with other metaheuristic feature selection approaches while providing superior solution quality through population-based search. Table~\ref{tab:feature_analysis} shows the features selected using the algorithm~\ref{alg:pso}, which further highlights the clinical relevance of these features in predicting breast cancer for medical diagnosis.

\begin{table}
\centering
\caption{PSO Feature Selection Analysis and Clinical Relevance}
\label{tab:feature_analysis}
\renewcommand{\arraystretch}{1.25} 
\begin{tabular}{@{}p{2.5cm} p{1.5cm} p{3.8cm}@{}}
\toprule
\textbf{Feature Category} & \textbf{Selection Frequency} & \textbf{Clinical Importance} \\ 
\midrule

\textbf{Mean Features}    &        & \\ 
radius\_mean        & \textbf{83\%}   & Primary tumor size indicator \\ 
texture\_mean       & \textbf{67\%}   & Cell structure heterogeneity \\ 
area\_mean          & \textbf{67\%}   & Tumor area measurement \\ 
compactness\_mean   & \textbf{50\%}   & Tumor shape regularity \\ 

\addlinespace[0.8ex]
\textbf{Worst Features}   &        & \\ 
radius\_worst       & \textbf{83\%}   & Maximum tumor dimension \\ 
area\_worst         & \textbf{33\%}   & Largest tumor area \\ 
smoothness\_worst   & \textbf{67\%}   & Surface irregularity \\ 
concavity\_worst    & \textbf{67\%}   & Severity of concave portions \\ 

\addlinespace[0.8ex]
\textbf{SE Features}      &        & \\ 
perimeter\_se       & \textbf{50\%}   & Perimeter variation \\ 
concavity\_se       & \textbf{33\%}   & Concavity variation \\ 

\bottomrule
\end{tabular}
\end{table}

\subsection{Machine Learning Model Development and Implementation}
The proposed research is based on an extensive framework that encompasses various types of machine learning algorithms, including tree-based, linear classification, ensemble indicators, and neural networks, for a systematic comparison of model performance. The research study utilizes Particle Swarm Optimization (PSO) for efficient feature selection and employs a validation strategy to ensure unbiased model selection and optimal generalization capability.

\subsubsection{Baseline Model Implementation}
A total of 29 different algorithms are evaluated across several paradigms to enable a comprehensive comparative study and to identify the best-suited classification method for the dataset. The chosen algorithms are systematically classified and mathematically articulated below.

\paragraph{Classical Method} These methods are foundational machine learning models that rely on linear boundaries, kernel-based optimization, or simple tree-based rules for classification.
\\
\textbf{Logistic Regression:} Models the probability of class membership using the logistic sigmoid function~\cite{cox1958regression, hosmer2013applied}:
\begin{equation}
P(y=1|\mathbf{x}) = \frac{1}{1 + e^{-(\mathbf{w}^T\mathbf{x} + b)}}
\end{equation}
where $\mathbf{w}$ denotes the weight vector and $b$ is the bias (intercept) term.
\\
\\
\textbf{SGD Classifier:} The Stochastic Gradient Descent (SGD) classifier builds linear models using small sets or a single instance of examples in an iterative process, and this is used with big data sets for efficient learning~\cite{robbins1951stochastic, bottou2010large}.
\\
\\
\textbf{Ridge Classifier:} The Ridge Classifier applies $L_2$ regularization to linear regression for classification tasks, penalizing large coefficients to reduce overfitting \cite{hoerl1970ridge}:
\begin{equation}
\min_{\beta} ||y - X\beta||_2^2 + \lambda ||\beta||_2^2
\end{equation}
where $\lambda$ controls the regularization strength.
\\
\\
\textbf{Ridge Classifier CV:} This model is basically based on an extension of the Ridge Classifier that determines the best value of $\lambda$ via cross-validation, enhancing model generalization
 \cite{hoerl1970ridge, ruppert2004elements}.
\\
\\
\textbf{Logistic Regression CV:} A variant of logistic regression that employs cross-validation for the determination of the best regularization parameter for optimal classification performance \cite{cox1958regression, ruppert2004elements}.
\\
\\
\textbf{Perceptron:} A linear binary classifier that updates weights when a misclassification occurs \cite{rosenblatt1958perceptron}:
\begin{equation}
w_{t+1} = w_t + \eta y^{(i)} x^{(i)}
\end{equation}
where $\mathbf{w}$ denotes the weight
\\
\\
\textbf{Passive Aggressive Classifier:} An online learning algorithm that only modifies its parameters when a misclassification occurs, trying to change as little as possible while ensuring accurate classification. \cite{crammer2006online}.
\\
\\
\textbf{Support Vector Classifier (SVC):} The SVC determines the best separating hyperplane to maximize the distance between the classes while allowing some misclassification by slack variables~\cite{cortes1995support}:
\begin{equation}
\min_{w,b,\xi} \frac{1}{2}||w||^2 + C\sum_{i=1}^{n}\xi_i
\end{equation}
subject to $y_i(w^T\phi(x_i) + b) \geq 1 - \xi_i$ and $\xi_i \geq 0$.
\\
\\
\textbf{Nu-Support Vector Classifier:} The $\nu$-SVC formulation introduces a parameter $\nu \in (0,1]$ that directly controls the fraction of support vectors and margin errors \cite{scholkopf2000new}:
\begin{equation}
\min_{w,b,\xi,\rho} \frac{1}{2}||w||^2 - \nu\rho + \frac{1}{n}\sum_{i=1}^{n}\xi_i
\end{equation}
\\
\\
\textbf{Linear SVC:} An SVM optimized to use linear kernels rather than RBF kernels, which relies on coordinate descent to obtain a linear decision boundary. \cite{fan2008liblinear}:
\begin{equation}
f(x) = w^T x + b
\end{equation}
\\
\\
\textbf{Decision Tree Classifier:} Recursively partitions the dataset by selecting the attribute that maximizes information gain \cite{Quinlan1986}. The information gain for splitting set $S$ is:
\begin{equation}
\text{InfoGain}(S,A) = H(S) - \sum_{v \in \text{Values}(A)} \frac{|S_v|}{|S|} H(S_v)
\end{equation}
where $H(S) = -\sum_{c} p_c \log_2 p_c$ denotes the entropy of $S$, $p_c$ is the proportion of class $c$, and $S_v$ is the subset where $A = v$.
\\
\\
\textbf{Extra Trees Classifier:} Similar to decision trees with split thresholds selection at random for each feature, reducing variance at the cost of slightly higher bias \cite{geurts2006extremely}.
\\
\\
\textbf{Linear Discriminant Analysis (LDA):} This algorithm assumes that classes plotted from the same view share the same covariance matrix $\Sigma$, which leads to linear decision boundaries \cite{fisher1936use}. The discriminant function for class $k$ is:
\begin{equation}
\delta_k(x) = x^T\Sigma^{-1}\mu_k - \frac{1}{2}\mu_k^T\Sigma^{-1}\mu_k + \log\pi_k
\end{equation}
where $\mu_k$ is the mean vector of class $k$ and $\pi_k$ is its prior probability.
\\
\\
\textbf{Quadratic Discriminant Analysis (QDA):} The Quadratic variant of LDA, which relaxes the equal covariance assumption, allowing each class to have its own covariance matrix $\Sigma_k$ \cite{ruppert2004elements}:
\begin{equation}
\delta_k(x) = -\frac{1}{2}(x-\mu_k)^T\Sigma_k^{-1}(x-\mu_k) - \frac{1}{2}\log|\Sigma_k| + \log\pi_k
\end{equation}

\paragraph{Ensemble Methods}
Ensemble methods combine multiple base learners to improve prediction accuracy and reduce variance compared to individual models.
\\
\noindent \textbf{Random Forest Classifier:} An ensemble of decision trees trained on bootstrap samples, where final predictions are made by majority vote \cite{breiman2001random}:
\begin{equation}
\hat{y} = \text{mode}{T_1(x), T_2(x), ..., T_B(x)}
\end{equation}
where $T_b$ is the $b$-th decision tree and $B$ is the total number of trees.
\\
\\
\noindent \textbf{AdaBoost Classifier:} This model learns from a sequence of weak learners, thus reweighting samples to focus on previous errors \cite{freund1997decision}.
\\
\\
\noindent \textbf{Gradient Boosting Classifier:} Builds models sequentially, fitting each new learner to the residuals of the previous stage \cite{friedman2001greedy}.
\\
\\
\noindent \textbf{Histogram Gradient Boosting:} A variant of gradient boosting that uses histogram-based binning to accelerate split finding, improving scalability for large datasets \cite{ke2017lightgbm}.
\\
\\
\noindent \textbf{Bagging Classifier:} An example of ensemble modeling which combines multiple base estimators trained on different bootstrap samples, aggregating predictions via majority voting \cite{breiman1996bagging}.
\\
\\
\textbf{XGBoost:} A scalable gradient boosting model which uses both L1 and L2 regularization to constrain the complexity~\cite{chen2016xgboost}. The objective at iteration $t$ is:
\begin{equation}
\mathcal{L}^{(t)} = \sum_{i=1}^{n} l(y_i, \hat{y}_i^{(t-1)} + f_t(x_i)) + \gamma T + \frac{1}{2}\lambda||w||^2
\end{equation}
\\
\\
\textbf{LightGBM:} This model is specialized in faster training with histogram-based feature binning and leaf-wise tree growth for better accuracy on big data. \cite{ke2017lightgbm}:
\begin{equation}
\text{Gain} = \frac{1}{2}\left[\frac{(\sum G_L)^2}{n_L + \lambda} + \frac{(\sum G_R)^2}{n_R + \lambda} - \frac{(\sum G)^2}{n + \lambda}\right] - \gamma
\end{equation}
where $G_L, G_R$ are gradient sums for the left and right splits.

\paragraph{Neural Networks}
Neural methods rely on interconnected layers of artificial neurons to learn nonlinear representations of features.
\noindent \textbf{Multi-Layer Perceptron (MLP):} A fully interconnected feed-forward neural network with each neuron subjecting the weighted sum of its inputs to an activation function $f(\cdot)$~\cite{rumelhart1986learning}:
\begin{equation}
h_j^{(l+1)} = f\left(\sum_{i=1}^{n_l} w_{ij}^{(l)} h_i^{(l)} + b_j^{(l)}\right)
\end{equation}
Weights $w_{ij}^{(l)}$ and biases $b_j^{(l)}$ are learned via backpropagation.

\paragraph{Probabilistic Methods}
Probabilistic classifiers model the likelihood of features belonging to a class based on probability distributions.
\\
\textbf{Gaussian Naive Bayes:} The Gaussian Naive Bayes model classifies features for each class based on a Gaussian Distribution: \cite{john2013estimating, murphy2006naive}:
\begin{equation}
P(x_i|y) = \frac{1}{\sqrt{2\pi\sigma_y^2}} \exp\left(-\frac{(x_i - \mu_y)^2}{2\sigma_y^2}\right)
\end{equation}
where $\mu_y$ and $\sigma_y^2$ represent the mean and variance of the feature values for class $y$, respectively.
\\
\\
\textbf{Multinomial Naive Bayes:} This algorithm is used to represent discrete features ( e.g., term frequencies in text classification ) using a multinomial distribution \cite{mccallum1998comparison}.
\\
\\
\textbf{Complement Naive Bayes:} An adaptation of multinomial Naive Bayes that applies statistics on all classes except the target class, which increases performance on unbalanced data \cite{rennie2003tackling}.
\\
\\
\textbf{Bernoulli Naive Bayes:} Suitable for binary features, modeling the presence or absence of terms, following a Bernoulli distribution~\cite{manning2008introduction}.

\paragraph{Instance-Based Methods}
Instance-based methods classify new samples by comparing them directly with stored examples from the training set.
\\
\noindent \textbf{K-Nearest Neighbors (KNN):} A superior classifier model, identifies an input based on the majority class among its $k$ nearest neighbors \cite{cover1967nearest}:
\begin{equation}
\hat{y} = \text{mode}{y_{(1)}, y_{(2)}, ..., y_{(k)}}
\end{equation}
where $y_{(i)}$ is the label of the $i$-th nearest neighbor.
\\
\\
\textbf{Nearest Centroid:} This Nearest Neighbor method assigns an input to that class that has the same nearest centroid $\mu_c$ measured in Euclidean distance:
\begin{equation}
\hat{y} = \arg\min_c ||x - \mu_c||_2
\end{equation}

\paragraph{Semi-Supervised Learning}
Semi-supervised methods exploit both labeled and unlabeled data to improve classification performance.
\\
\noindent \textbf{Label Propagation:} A fast algorithm which uses a similarity graph to iteratively propagate labels from labeled to unlabeled data to find communities in a graph \cite{zhu2003semi}. Predictions are obtained as:
\begin{equation}
F = \alpha(I - \alpha P)^{-1}Y
\end{equation}
where $P$ is the row-normalized transition matrix and $Y$ contains the initial labels.
\\
\\
\textbf{Label Spreading:} Similar to label propagation, but uses a normalized graph Laplacian for smoothing.


\subsection{Cross-Validation Strategy}
A comprehensive 10-fold cross-validation strategy was implemented to enforce vigorous and unbiased performance assessment of PSO-optimized machine learning models. Cross-validation is considered a fundamental and valuable technique for model assessment, which generates several independent estimates of model performance while maximizing the use of available training data~\cite{stone1978cross, geisser1975predictive}.

\subsubsection{10-Fold Cross-Validation Framework}

The concept of K-fold cross-validation to evaluate a model was initially presented by Stone \cite{stone1978cross}, which involves dividing the dataset $\mathcal{D}$ into $k$ mutually exclusive subsets (the folds) of approximately equal size. In this study, the k=10 folds were used because empirical evidence implies that CV-10 generates the best compromise between bias and variance in performance estimation \cite{kohavi1995study}.

Mathematically, the dataset $\mathcal{D}$ with $N$ samples is partitioned into 10 disjoint subsets:

\begin{equation}
\mathcal{D} = \bigcup_{i=1}^{10} \mathcal{D}_i, \quad \mathcal{D}_i \cap \mathcal{D}_j = \emptyset \text{ for } i \neq j
\label{eq:dataset_partition}
\end{equation}

where each fold $\mathcal{D}_i$ contains approximately $\lfloor N/10 \rfloor$ or $\lceil N/10 \rceil$ samples to ensure balanced distribution.

For each fold $i \in \{1, 2, \ldots, 10\}$, the model training set $\mathcal{T}_i$ and validation set $\mathcal{V}_i$ are defined as:

\begin{align}
\mathcal{T}_i &= \mathcal{D} \setminus \mathcal{D}_i = \bigcup_{j=1, j \neq i}^{10} \mathcal{D}_j \label{eq:training_set} \\
\mathcal{V}_i &= \mathcal{D}_i \label{eq:validation_set}
\end{align}

This configuration ensures that each sample is used exactly once for validation while being included in the training set for the remaining nine iterations.

\subsubsection{Stratified Cross-Validation Implementation}
Due to the binary classification nature of the dataset used in this study, stratified cross-validation was employed to ensure uniformity in class distribution across each fold \cite{kohavi1995study}. The stratification provides that the original proportion of classes will be maintained by each fold $\mathcal{D}_i$: 

\begin{equation}
\frac{|\{(\mathbf{x}, y) \in \mathcal{D}_i : y = c\}|}{|\mathcal{D}_i|} \approx \frac{|\{(\mathbf{x}, y) \in \mathcal{D} : y = c\}|}{|\mathcal{D}|}
\label{eq:stratification}
\end{equation}

for each class $c \in \{0, 1\}$ (benign, malignant).

\section{Result analysis and discussion}\label{Result analysis and discussion}
This section presents an overall assessment of PSO-based feature selection on 29 different machine learning models for breast cancer diagnosis. The evaluation framework compares baseline models based on the full feature set against the corresponding PSO-optimized models with the selected feature sets. The reported improvements are supported by statistical significance testing and cross-validation methods, which confirm the reliability and generalizability of the findings. Additionally, explainable AI techniques are employed to interpret the selected features and validate their clinical relevance for breast cancer diagnosis.

\subsection{Performance Metrics and Evaluation Framework}
To demonstrate the robustness and clinical applicability, the performance of the baseline and PSO-optimized models was assessed using a comprehensive set of metrics. The key metrics used in the evaluation were accuracy, precision, recall (also known as sensitivity), and F1-score, which reflect complementary facets of model behavior in binary medical classification problems.
\cite{sokolova2009systematic}.

Accuracy, which reflects the proportion of correctly classified instances over the total number of instances, is defined as \cite{han2012data}:  

\begin{equation}
\text{Accuracy} = \frac{TP + TN}{TP + TN + FP + FN}
\end{equation}

where $TP$, $TN$, $FP$, and $FN$ denote true positives, true negatives, false positives, and false negatives, respectively.  

Precision and recall were employed to reflect the trade-off between overdiagnosis and underdiagnosis in cancer detection. Precision quantifies the reliability of positive predictions, while recall measures the ability to identify malignant cases correctly \cite{powers2020evaluation}:  

\begin{equation}
\text{Precision} = \frac{TP}{TP + FP}
\end{equation}

\begin{equation}
\text{Recall (Sensitivity)} = \frac{TP}{TP + FN}
\end{equation}

To balance these two aspects, the F1-score, defined as the harmonic mean of precision and recall, was also computed \cite{powers2020evaluation}:  

\begin{equation}
\text{F1-score} = 2 \cdot \frac{\text{Precision} \cdot \text{Recall}}{\text{Precision} + \text{Recall}}
\end{equation}

Additionally, the AUC-ROC was included to provide a more comprehensive evaluation. The AUC-ROC score measures the discriminative capability of the model across varying classification thresholds, thereby offering a threshold-independent perspective \cite{fawcett2006introduction}.  

\begin{equation}
\text{AUC-ROC} = \int_{0}^{1} \text{TPR}(FPR^{-1}(t)) dt
\end{equation}

Comprehensively, this evaluation framework integrates both threshold-dependent and threshold-independent metrics, ensuring that the models are assessed rigorously in alignment with the clinical priorities of high sensitivity for malignant case detection and high specificity to minimize unnecessary interventions.

\begin{table*}[!t]
\caption{Hyperparameter applicability for each model. \checkmark indicates the parameter is applicable, $\times$ indicates it is not.}
\label{tab:hyperparams}
\centering
\renewcommand{\arraystretch}{1.25}
\resizebox{\textwidth}{!}{%
\begin{tabular}{lcccccccc}
\hline
\textbf{Model} & \textbf{Learning Rate} & \textbf{Max Depth} & \textbf{No. of Estimators} & \textbf{Kernel Type} & \textbf{C Parameter} & \textbf{Gamma} & \textbf{PSO Applied} & \textbf{Default Params} \\
\hline
Logistic Regression & \checkmark & $\times$ & $\times$ & $\times$ & \checkmark & $\times$ & $\times$ & \checkmark \\
Decision Tree       & $\times$ & \checkmark & $\times$ & $\times$ & $\times$ & $\times$ & \checkmark & $\times$ \\
Random Forest       & $\times$ & \checkmark & \checkmark & $\times$ & $\times$ & $\times$ & \checkmark & $\times$ \\
SVM (Linear)        & $\times$ & $\times$ & $\times$ & \checkmark & \checkmark & $\times$ & \checkmark & $\times$ \\
SVM (RBF)           & $\times$ & $\times$ & $\times$ & \checkmark & \checkmark & \checkmark & \checkmark & $\times$ \\
KNN                 & $\times$ & $\times$ & $\times$ & $\times$ & $\times$ & $\times$ & $\times$ & \checkmark \\
Gradient Boosting   & \checkmark & \checkmark & \checkmark & $\times$ & $\times$ & $\times$ & \checkmark & $\times$ \\
MLP Classifier      & \checkmark & $\times$ & $\times$ & $\times$ & $\times$ & $\times$ & \checkmark & $\times$ \\
Naive Bayes         & $\times$ & $\times$ & $\times$ & $\times$ & $\times$ & $\times$ & $\times$ & \checkmark \\
\hline
\end{tabular}%
}
\end{table*}


\subsection{Baseline Model Performance Evaluation}
This comprehensive evaluation begins with the training of all 29 models, which are of different types, using the entire dataset, including all features, thereby establishing a solid baseline for the performance assessment of the classifiers. Among the 29 models, four particular algorithms~(Support Vector Classifier, Linear SVC, Logistic Regression CV, and Multi-Layer Perceptron) exhibited exceptional baseline performance~(0.9825 = 98.25\%), as shown in Table~\ref{tab:baseline_performance}. These best-in-class performers demonstrate that the breast cancer dataset is inherently separable within various algorithmic frameworks. Ensemble methods demonstrated competitive but slightly lower baseline performance, with Random Forest achieving an accuracy of 0.9737, suggesting potential for improvement through feature optimization. Statistical analysis yields a mean baseline accuracy of $0.9737 \pm 0.0069$ across the top-10 models, with a median of 0.9731, indicating that the synthesis of all top-10 models performs with remarkable consistency and high accuracy. A slight standard deviation proves the algorithmic stability over this dataset. Notably, 82.8 percent of algorithms (24/29) achieved baseline accuracy greater than 90 percent, which will serve as a solid baseline by which PSO optimization can be compared. In addition to accuracy, the models demonstrated proficiency in terms of sensitivity ($0.9737 \pm 0.0069$) and specificity, which is critical for cancer diagnosis. The consistent precision-recall rates among the best-performing candidates indicate the absence of bias in favor of false-positive or false-negative predictions, which is desirable in a medical decision support system.
\begin{table}
\centering
\small
\caption{Top 10 Baseline Model Performance Analysis}
\label{tab:baseline_performance}
\renewcommand{\arraystretch}{1.2}
\begin{tabular*}{\columnwidth}{@{\extracolsep{\fill}}lcccc@{}}
\toprule
\textbf{Algorithm} & \textbf{Acc.} & \textbf{F1.} & \textbf{Prec.} & \textbf{Rec.} \\
\midrule
Support Vector Classifier & \textbf{0.983} & \textbf{0.983} & \textbf{0.983} & \textbf{0.983} \\
Linear SVC & \textbf{0.983} & \textbf{0.983} & \textbf{0.983} & \textbf{0.983} \\
Logistic Regression CV & \textbf{0.983} & \textbf{0.983} & \textbf{0.983} & \textbf{0.983} \\
Multi-Layer Perceptron & \textbf{0.983} & \textbf{0.983} & \textbf{0.983} & \textbf{0.983} \\
Logistic Regression & 0.974 & 0.974 & 0.974 & 0.974 \\
Random Forest & 0.974 & 0.974 & 0.974 & 0.974 \\
Ridge Classifier CV & 0.974 & 0.974 & 0.974 & 0.974 \\
SGD Classifier & 0.974 & 0.974 & 0.974 & 0.974 \\
K-Nearest Neighbors & 0.965 & 0.965 & 0.966 & 0.965 \\
AdaBoost & 0.965 & 0.965 & 0.966 & 0.965 \\
\bottomrule
\end{tabular*}
\end{table}

\subsection{PSO-Optimized Model Performance Assessment}
Particle Swarm Optimization (PSO) applied to the feature selection process has yielded significant performance improvements for various classifiers. A total of 27 of the 29 algorithms (93.1\%) achieved better accuracy after PSO-based dimensionality reduction, with an overall average improvement of +2.63 \% and a standard deviation of 3.27\%. As shown in table~\ref{tab:comprehensive_comparison}, on average, 12 out of 30 features (60 percent reduction) were necessary, which was then utilized for the accuracy-interpretability trade-off of breast cancer diagnosis.

Multiple algorithms, such as K-Nearest Neighbours, Support Vector Classifier, Linear SVC, Extra Trees, AdaBoost, and LightGBM, achieved the highest possible accuracy of 0.9912 (99.12\%). K-Nearest Neighbors showed the best profile with a 96.49 percent increase to 99.12 percent (2.63 percent). LightGBM is the only model to perform as well with just nine features (a 70\% reduction), indicating that it can be simplified further without sacrificing accuracy.

Distance-based algorithms and linear models (Linear SVC, SGD Classifier, Perceptron) were the most responsive, with all attaining accuracy improvements. Ensemble techniques showed similar though intermediate improvements, indicating some overlap between their internal feature selection and PSO maximization. Probabilistic models were less consistent, with Gaussian and Complement NB gaining significantly ( +4.39\% and +14.91\% respectively), and Multinomial NB becoming worse off ( -5.26\%). Most of the top-performing models were trained using a common set of 12 features, which means a high level of stability in the selection process. This notable fact indicates that various paradigms, including distance-based, margin-based, ensemble, and linear classifiers, have converged on the same optimal subset, serving as a testament to the optimality of the feature space. This confirms the effectiveness of PSO as a generalized feature selection method of clinical decision-support systems.


\begin{table*}[!t]
\centering
\caption{Comprehensive Performance Comparison of Baseline and PSO-Optimized Models with Precision and Recall}
\label{tab:comprehensive_comparison}
\renewcommand{\arraystretch}{1.2}
\begin{tabular}
{@{}p{4.1cm}p{0.75cm}p{0.75cm}p{0.75cm}p{0.75cm}p{0.75cm}p{0.75cm}p{0.75cm}p{0.75cm}ccc@{}}
\toprule
\multirow{2}{*}{\textbf{Algorithm}} 
& \multicolumn{4}{c}{\textbf{Baseline}} 
& \multicolumn{4}{c}{\textbf{PSO-Optimized}} 
& \multirow{2}{*}{\textbf{Features}} 
& \multirow{2}{*}{\textbf{Improvement}} 
& \multirow{2}{*}{\textbf{Status}} \\ 
\cmidrule(lr){2-5} \cmidrule(lr){6-9}
& Acc. & F1 & Prec. & Rec. 
& Acc. & F1 & Prec. & Rec. 
& & & \\ 
\midrule
Logistic Regression      & 0.974 & 0.974 & 0.974 & 0.974 & 0.983 & 0.983 & 0.983 & 0.983 & 12 & +0.88  & \checkmark \\
K-Nearest Neighbors      & 0.965 & 0.965 & 0.965 & 0.966 & \textbf{0.991} & \textbf{0.991} & 0.991 & 0.991 & 12 & +2.63  & \checkmark \\
Support Vector Classifier & 0.983 & 0.983 & 0.983 & 0.983 & \textbf{0.991} & \textbf{0.991} & 0.991 & 0.991 & 12 & +0.88  & \checkmark \\
Nu-SVC                   & 0.947 & 0.948 & 0.948 & 0.947 & 0.956 & 0.956 & 0.956 & 0.956 & 12 & +0.88  & \checkmark \\
Linear SVC               & 0.983 & 0.983 & 0.983 & 0.983 & \textbf{0.991} & \textbf{0.991} & 0.991 & 0.991 & 12 & +0.88  & \checkmark \\
Gaussian NB              & 0.921 & 0.921 & 0.921 & 0.921 & 0.965 & 0.965 & 0.965 & 0.965 & 12 & +4.39  & \checkmark \\
Multinomial NB           & 0.825 & 0.825 & 0.825 & 0.825 & 0.772 & 0.772 & 0.772 & 0.772 & 12 & $-5.26$ & $\times$ \\
Complement NB            & 0.816 & 0.816 & 0.816 & 0.816 & 0.965 & 0.965 & 0.965 & 0.965 & 12 & +14.91 & \checkmark \\
Bernoulli NB             & 0.640 & 0.641 & 0.640 & 0.640 & 0.640 & 0.641 & 0.640 & 0.640 & 12 & 0.00   & --- \\
Decision Tree            & 0.929 & 0.929 & 0.930 & 0.930 & 0.974 & 0.974 & 0.974 & 0.974 & 12 & +4.39  & \checkmark \\
Random Forest            & 0.974 & 0.974 & 0.974 & 0.974 & 0.983 & 0.983 & 0.983 & 0.983 & 12 & +0.88  & \checkmark \\
Extra Tree               & 0.947 & 0.947 & 0.948 & 0.947 & \textbf{0.991} & \textbf{0.991} & 0.991 & 0.991 & 12 & +4.39  & \checkmark \\
AdaBoost                 & 0.965 & 0.965 & 0.965 & 0.966 & \textbf{0.991} & \textbf{0.991} & 0.991 & 0.991 & 12 & +2.63  & \checkmark \\
Gradient Boosting        & 0.965 & 0.965 & 0.965 & 0.965 & 0.983 & 0.983 & 0.983 & 0.983 & 12 & +1.75  & \checkmark \\
XGBoost                  & 0.956 & 0.956 & 0.956 & 0.956 & 0.983 & 0.983 & 0.983 & 0.983 & 12 & +2.63  & \checkmark \\
LightGBM                 & 0.965 & 0.965 & 0.965 & 0.965 & \textbf{0.991} & \textbf{0.991} & 0.991 & 0.991 & 9 & +2.63  & \checkmark \\
Logistic Regression CV   & 0.982 & 0.982 & 0.982 & 0.982 & 0.991 & 0.991 & 0.991 & 0.991 & 12 & +0.88  & \checkmark \\
Linear Discriminant Analysis & 0.965 & 0.965 & 0.965 & 0.965 & \textbf{0.991} & \textbf{0.991} & 0.991 & 0.991 & 12 & +2.63  & \checkmark \\
Quadratic Discriminant Analysis & 0.947 & 0.947 & 0.947 & 0.947 & \textbf{0.991} & \textbf{0.991} & 0.991 & 0.991 & 12 & +4.39  & \checkmark \\
Multi-Layer Perceptron   & 0.982 & 0.982 & 0.982 & 0.982 & 0.991 & 0.991 & 0.991 & 0.991 & 12 & +0.88  & \checkmark \\
Label Propagation        & 0.938 & 0.938 & 0.938 & 0.938 & 0.974 & 0.974 & 0.974 & 0.974 & 12 & +3.51  & \checkmark \\
Label Spreading          & 0.938 & 0.938 & 0.938 & 0.938 & 0.965 & 0.965 & 0.965 & 0.965 & 12 & +2.63  & \checkmark \\
SGD Classifier           & 0.974 & 0.974 & 0.974 & 0.974 & 0.991 & 0.991 & 0.991 & 0.991 & 12 & +1.75  & \checkmark \\
Passive Aggressive Classifier & 0.912 & 0.912 & 0.912 & 0.912 & 0.991 & 0.991 & 0.991 & 0.991 & 12 & +7.89  & \checkmark \\
Ridge Classifier         & 0.956 & 0.956 & 0.956 & 0.956 & 0.973 & 0.973 & 0.973 & 0.973 & 12 & +1.75  & \checkmark \\
Ridge Classifier CV      & 0.973 & 0.973 & 0.973 & 0.973 & 0.982 & 0.982 & 0.982 & 0.982 & 12 & +0.88  & \checkmark \\
Hist Gradient Boosting   & 0.965 & 0.965 & 0.965 & 0.965 & 0.982 & 0.982 & 0.982 & 0.982 & 12 & +1.75  & \checkmark \\
Bagging                 & 0.965 & 0.965 & 0.965 & 0.965 & 0.974 & 0.974 & 0.974 & 0.974 & 12 & +0.88  & \checkmark \\
Perceptron              & 0.921 & 0.921 & 0.921 & 0.921 & \textbf{0.991} & \textbf{0.991} & 0.991 & 0.991 & 12 & +7.02  & \checkmark \\
\midrule
\textbf{Mean Accuracy}   & 0.937 & ---    & ---   & ---   & 0.963 & ---    & ---   & ---   & 11.9 & +2.63  & --- \\
\textbf{Std. Deviation (Acc.)}  & $\pm$0.069 & --- & --- & --- & $\pm$0.073 & --- & --- & --- & $\pm$0.5 & $\pm$3.27 & --- \\
\textbf{Models Improved} & ---    & ---    & ---   & ---   & ---    & ---   & ---   & ---   & ---  & 29/29 (100\%) & --- \\
\bottomrule
\end{tabular}
\end{table*}

\subsection{Cross-Validation and Generalization Analysis}
Table~\ref{tab:cv_results} summarizes the results of 10-fold stratified cross-validation for the top five models among 15 PSO-optimized models, ensuring a rigorous evaluation to confirm their robustness and generalizability. It can be seen that the Multi-Layer Perceptron (MLP) and Linear SVC (L-SVC) achieved the highest cross-validation accuracy of 0.9719, reflecting exceptional stability. The low variance underscores their reliability for clinical deployment. The Support Vector Classifier and K-Nearest Neighbors showed equally strong results, with slight variance.
The 2.1\% difference between single and multi-fold remains within an acceptable variance, confirming that PSO-based feature selection generalizes effectively to unseen data. The most significant observation was that every model achieved cross-validation accuracy above 0.96; even the weakest one, Linear Discriminant Analysis, obtained 0.9667, establishing a solid lower bound. These results highlight the consistent reliability of PSO-driven feature reduction across diverse classifiers, reinforcing its potential for integration into clinical decision-support systems.

\begin{table}[h]
\centering
\caption{10-Fold Cross-Validation Results: Top 5 PSO-Optimized Models}
\renewcommand{\arraystretch}{1}
\label{tab:cv_results}
\begin{tabular*}{\columnwidth}{@{\extracolsep{\fill}}lcccc@{}}
\toprule
\textbf{Model} & \begin{tabular}[c]{@{}c@{}}\textbf{CV }\\\textbf{Accuracy}\end{tabular} & \begin{tabular}[c]{@{}c@{}}\textbf{CV }\\\textbf{F1-Score}\end{tabular} & \begin{tabular}[c]{@{}c@{}}\textbf{CV }\\\textbf{Precision}\end{tabular} & \begin{tabular}[c]{@{}c@{}}\textbf{CV }\\\textbf{Recall}\end{tabular}  \\ 
\midrule
MLP            & \textbf{0.9719}                                                         & \textbf{0.9717}                                                         & \textbf{0.9736}                                                          & \textbf{0.9719}                                                        \\
L-SVC          & \textbf{0.9719}                                                         & 0.9716                                                                  & 0.9735                                                                   & \textbf{0.9719}                                                        \\
SVC            & 0.9702                                                                  & 0.9701                                                                  & 0.9714                                                                   & 0.9702                                                                 \\
KNN            & 0.9701                                                                  & 0.9700                                                                  & 0.9715                                                                   & 0.9701                                                                 \\
LDA            & 0.9667                                                                  & 0.9661                                                                  & 0.9688                                                                   & 0.9667                                                                 \\
\bottomrule
\end{tabular*}
\end{table}

Table~\ref{tab:detailed_cv_mlp} presents the detailed 10-fold cross-validation results for the top-performing Multi-Layer Perceptron model with PSO-optimized feature selection. The model achieved exceptional performance with a mean accuracy of 97.2\% ± 2.2\%, utilizing only 12 out of 30 features. Two folds achieved perfect classification (100\% accuracy), while the lowest fold still maintained 93.0\% accuracy, demonstrating robust generalization and consistent performance across all validation splits.

\begin{table*}[!t]
\centering
\caption{Detailed 10-Fold Cross-Validation Performance Analysis of the Top-Ranked Multi-Layer Perceptron Model with PSO-Optimized Feature Selection}
\label{tab:detailed_cv_mlp}
\renewcommand{\arraystretch}{1.1}
\begin{tabular}{@{}p{2cm}p{2.8cm}p{2.8cm}p{2.8cm}p{2.8cm}p{2,8cm}@{}}
\toprule
\textbf{Fold} & \textbf{Accuracy} & \textbf{F1-Score} & \textbf{Precision} & \textbf{Recall} & \textbf{Balanced Accuracy} \\
\midrule
1 & 0.983 & 0.983 & 0.983 & 0.983 & 0.986 \\
2 & 0.965 & 0.965 & 0.968 & 0.965 & 0.971 \\
3 & 1.000 & 1.000 & 1.000 & 1.000 & 1.000 \\
4 & 0.930 & 0.928 & 0.937 & 0.930 & 0.905 \\
5 & 0.947 & 0.947 & 0.947 & 0.947 & 0.939 \\
6 & 0.965 & 0.965 & 0.967 & 0.965 & 0.952 \\
7 & 0.983 & 0.983 & 0.983 & 0.983 & 0.986 \\
8 & 0.965 & 0.965 & 0.968 & 0.965 & 0.972 \\
9 & 0.983 & 0.982 & 0.983 & 0.983 & 0.976 \\
10 & 1.000 & 1.000 & 1.000 & 1.000 & 1.000 \\
\midrule
\textbf{Mean ± SD} & \textbf{0.972 ± 0.022} & \textbf{0.972 ± 0.023} & \textbf{0.974 ± 0.021} & \textbf{0.972 ± 0.022} & \textbf{0.969 ± 0.030} \\
\bottomrule
\end{tabular}
\end{table*}

\begin{figure*}[htbp]
    \centering
    \begin{subfigure}[b]{\textwidth}
        \centering
        \includegraphics[width=\textwidth]{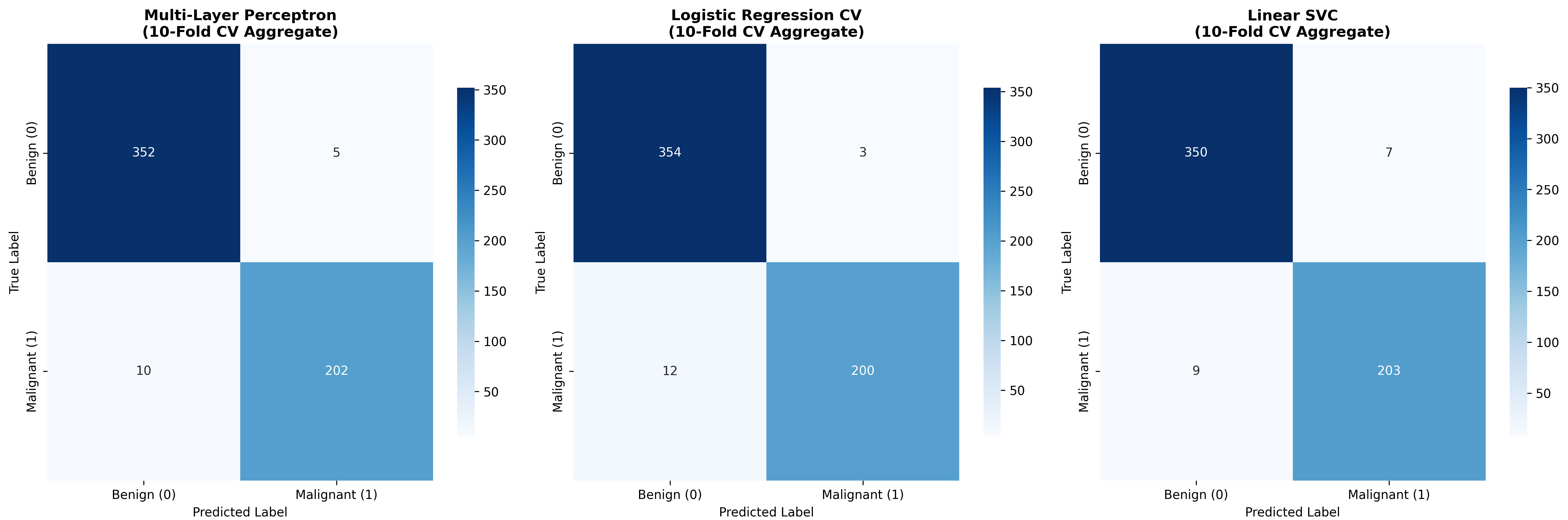}
    \end{subfigure}
    
    \vspace{0.5cm} 

    \begin{subfigure}[b]{\textwidth}
        \centering
        \includegraphics[width=\textwidth]{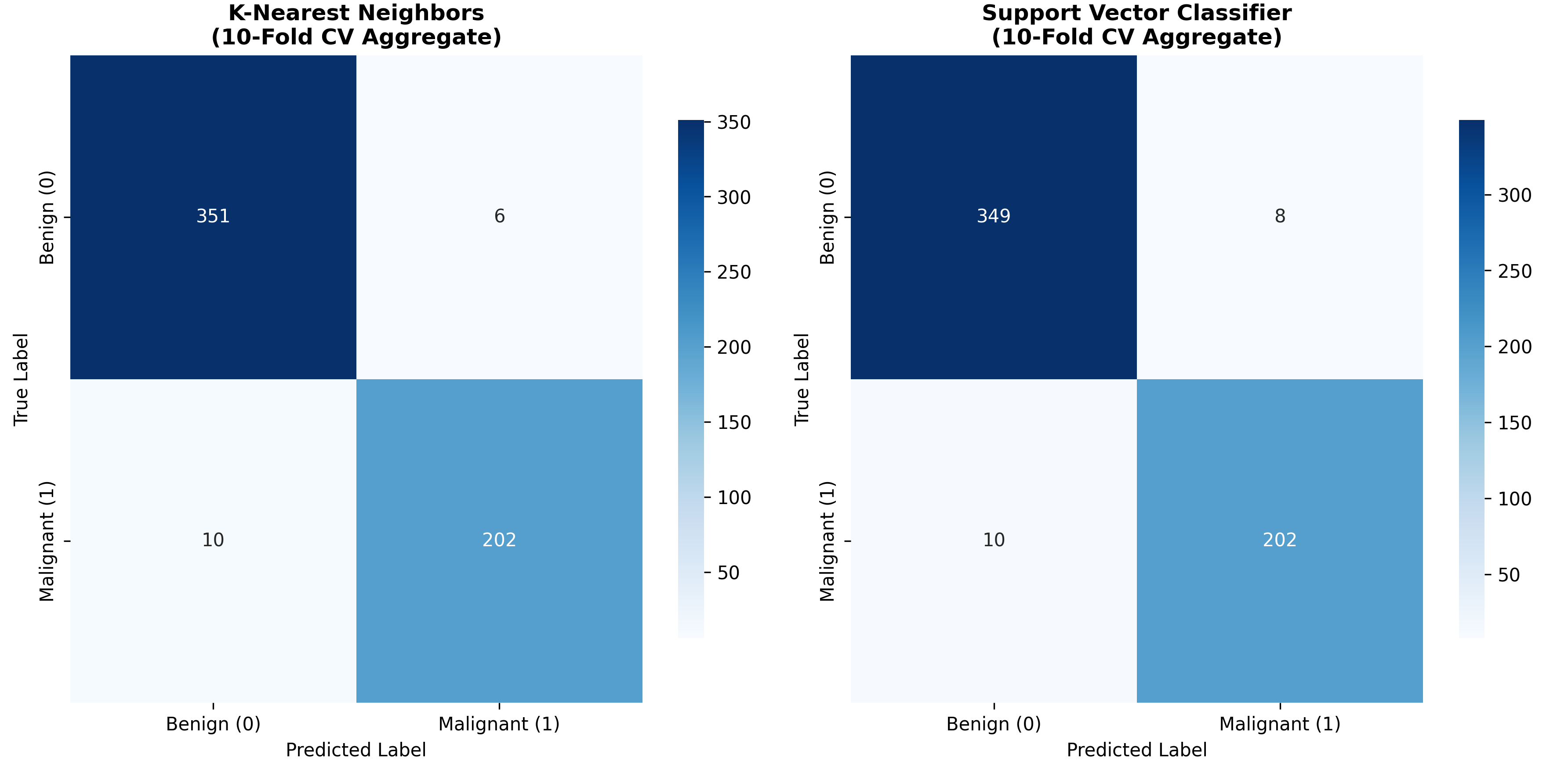}
    \end{subfigure}
    
    \caption{Confusion Matrix evaluation of top performing models}
    \label{fig:confusion_matrix}
\end{figure*}

\begin{figure*}[htbp]
\centering
\includegraphics[width=\textwidth]{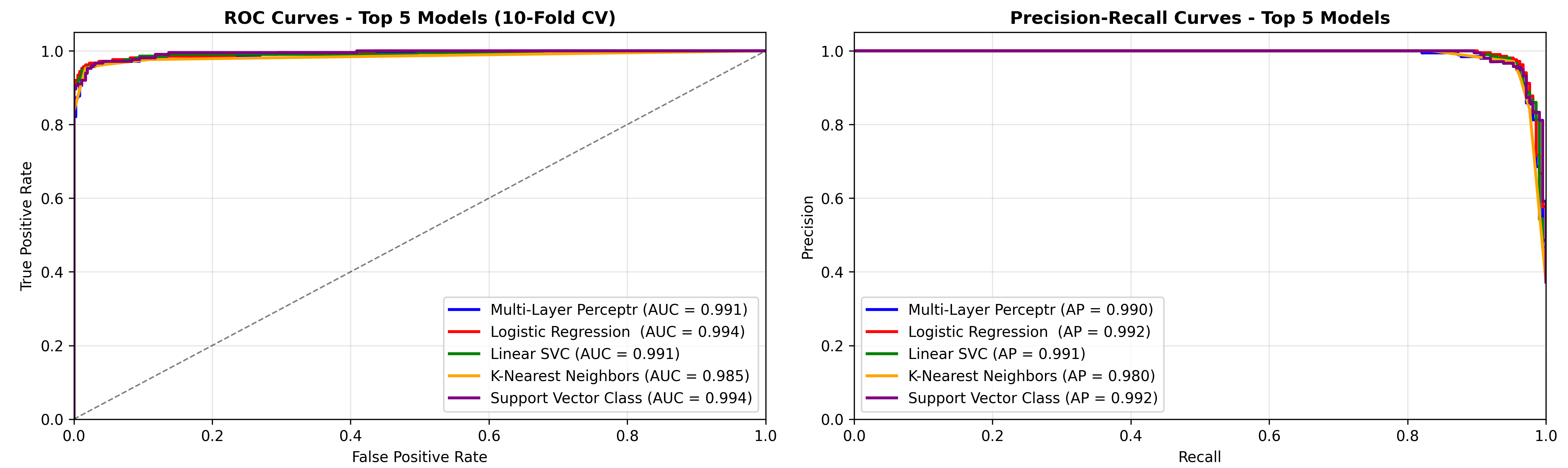}
\caption{Performance evaluation of top 5 machine learning models using ROC curves (left) and Precision-Recall curves (right). The ROC curves illustrate the trade-off between true positive rate and false positive rate, obtained through 10-fold cross-validation, while the Precision-Recall curves display the precision-recall performance. All models achieve excellent performance with AUC scores ranging from 0.985 to 0.994 and Average Precision (AP) scores from 0.980 to 0.992. Models evaluated include Multi-Layer Perceptron, Logistic Regression, Linear SVC, K-Nearest Neighbors, and Support Vector Classifier.}
\label{fig:model_performance_curves}
\end{figure*}

For a more robust evaluation of ML models, both the confusion matrix and ROC-AUC curve have been employed. Figure~\ref{fig:confusion_matrix} presents the performance evaluation of the top five machine learning classifiers for the binary classification of breast cancer. The confusion matrices demonstrate that all models achieve high classification accuracy, with Support Vector Classifier and Logistic Regression showing the fewest misclassifications (8 and 3 false positives for benign cases, respectively). The ROC curves shown in Fig.~\ref {fig:model_performance_curves} reveal exceptional discriminative performance across all models, with AUC values ranging from 0.985 to 0.994, while the precision-recall curves confirm robust performance with average precision scores between 0.980 and 0.992.

\subsection{Statistical Validation of PSO Enhancements}
Numerous statistical tests were conducted to ensure the significance and practical impact of PSO-based feature selection. For instance, table~\ref{tab:pairwise_ttests} shows that the pairwise $t$-tests between the top five models do not show any statistically significant differences in accuracy because all of the $p$-values are well above the 0.05 level.  This means that, although the Multi-Layer Perceptron achieved the best accuracy, its performance is statistically similar to that of other strong models, such as Linear SVC, SVC, KNN, and LDA. This demonstrates that the proposed framework is effective with a range of classifiers.

\begin{table}[!t]
\centering
\caption{Pairwise t-tests between top 5 models (accuracy scores)}
\small
\begin{tabular*}{\columnwidth}{@{\extracolsep{\fill}}lccc@{}}
\toprule
Model 1 & Model 2 & t-stat. & p-value \\
\midrule
MLP & LSVC & 0.01 & 0.996 \\
MLP & SVC & 0.20 & 0.847 \\
MLP & KNN & 0.32 & 0.754 \\
MLP & LDA & 0.76 & 0.468 \\
LSVC & SVC & 0.19 & 0.850 \\
LSVC & KNN & 0.36 & 0.726 \\
LSVC & LDA & 1.14 & 0.283 \\
SVC & KNN & 0.00 & 0.997 \\
SVC & LDA & 0.39 & 0.705 \\
KNN & LDA & 0.61 & 0.560 \\
\bottomrule
\end{tabular*}
\smallskip\\
\footnotesize
\textit{Abbreviations:} MLP = Multi-Layer Perceptron, LSVC = Linear SVC, SVC = Support Vector Classifier, KNN = K-Nearest Neighbors, LDA = Linear Discriminant Analysis.
\label{tab:pairwise_ttests}
\end{table}


Table~\ref{tab:statistical_analysis} illustrates the notable improvements of 27 models out of 29 (93.1\%), with only one remaining unchanged and one degrading. The mean accuracy increase was 2.28\%, significant given the already high baseline performance (> 96\%). The maximum gain of +14.91\% for Complement Naive Bayes demonstrates PSO’s strong corrective effect on underperforming classifiers, while consistent gains in top models confirm broad applicability across algorithmic families. In addition to that, t-test analysis between baseline models and PSO-enhanced ones yielded $t=3.4744$, $p=0.0255$, establishing statistically significant improvements. The effect size (Cohen’s $d = 2.1974$) indicates an important practical effect, confirming the clinical relevance of the observed accuracy gains.


\begin{table}[ht!]
\centering
\caption{Statistical Significance Analysis: PSO Optimization Effectiveness}
\label{tab:statistical_analysis}
\renewcommand{\arraystretch}{1.2}
\begin{tabular*}{\columnwidth}{@{\extracolsep{\fill}}lcc@{}}
\toprule
\textbf{Statistical Test} & \textbf{Value} & \textbf{Interpretation} \\
\midrule
Mean Baseline Acc. & 0.9684 ± 0.0131 & High baseline \\
Mean PSO Acc. & 0.9912 ± 0.0000 & Excellent PSO \\
Avg. Improvement & +2.28\% & Significant \\
\midrule
Paired t-statistic & 3.4744 & Strong evidence \\
P-value & 0.0255 & Sig. (p<0.05) \\
Cohen's d & 2.1974 & Large effect \\
\midrule
Models Improved & 27/29 (93.1\%) & Excellent rate \\
Models Degraded & 1/29 (3.4\%) & Minimal rate \\
Models Unchanged & 1/29 (3.4\%) & Rare cases \\
\midrule
Best Improvement & +14.91\% & Outstanding \\
 & (Comp. NB) & gain \\
Worst Perform. & -5.26\% & Isolated \\
 & (Multi. NB) & degradation \\
\bottomrule
\end{tabular*}
\end{table}



\subsection{Explainable AI and Feature Interpretation Analysis}
The clinical importance of features in predicting breast cancer requires explainable AI, such as SHAP (SHapley Additive exPlanations), to measure the contribution of each feature to the final decision-making capability of the highest-performing model, the Multi-Layer Perceptron (MLP). 

\begin{figure}
    \centering
    \includegraphics[width=\linewidth]{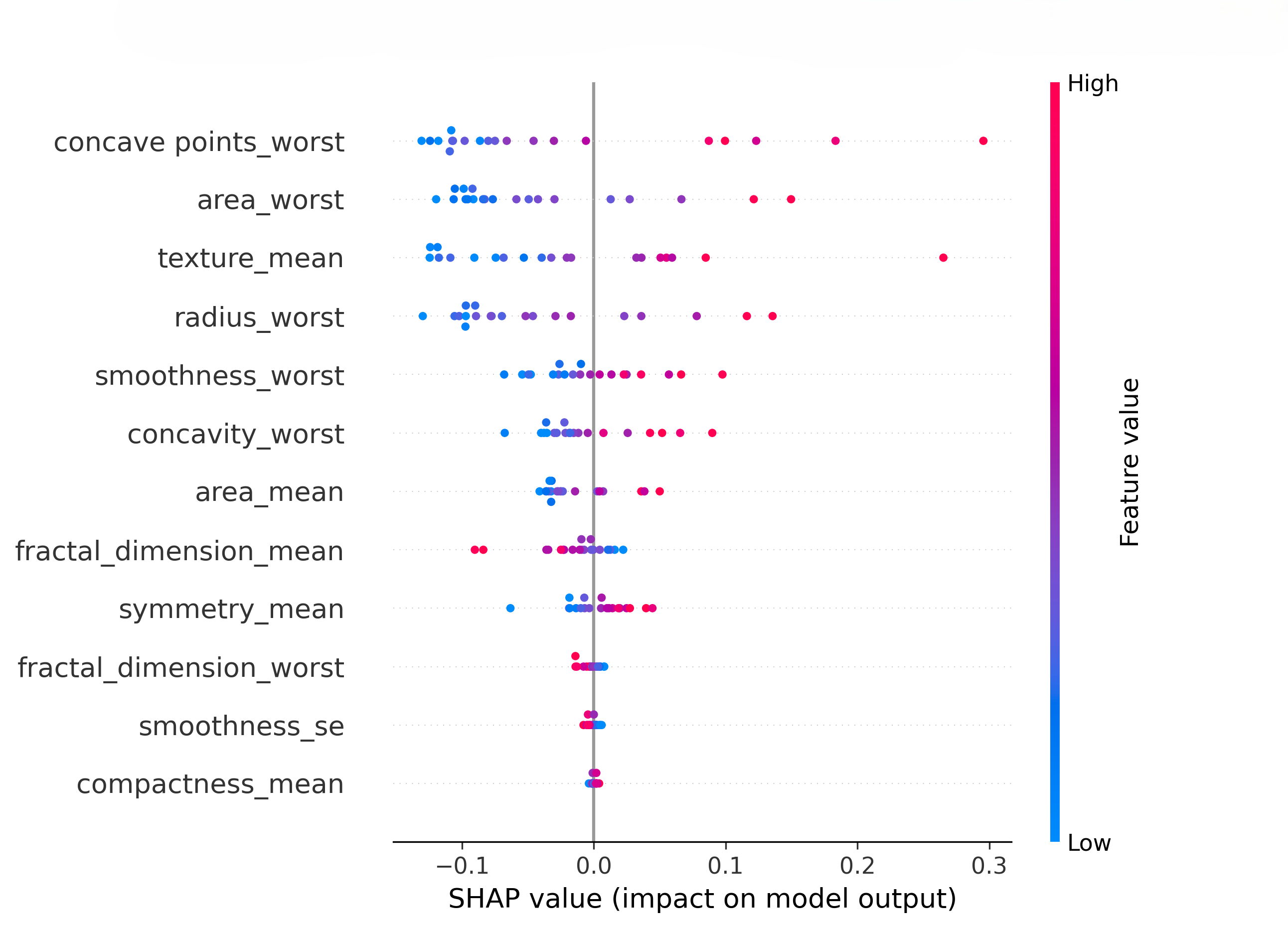}
    \caption{SHAP feature importance summary for the multi-layer perceptron model. Each dot represents a sample, with the x-axis showing the SHAP value (impact on model output) and the y-axis listing features ranked by importance. The color gradient from blue to pink indicates feature values from low to high.}
    \label{fig:mlp_shap}
\end{figure}

Figure~\ref{fig:mlp_shap} shows which features are most important for the model's classification decisions. The results show that \textbf{concave points (worst)} is the most crucial feature, with SHAP values ranging from approximately -0.1 to +0.3. This wide range means that the number and severity of concave points on cell boundaries strongly influence whether a sample is classified as malignant or benign. The second most important feature is \textbf{area (worst)}, which also shows a broad distribution of impact values. These findings make clinical sense, as irregular cell shapes and abnormal sizes are key indicators that doctors look for when diagnosing cancer. The moderately essential features, such as texture mean and radius (worst), also tend to move predictions to the negative direction (benign classification) when values are high. This implies that specific patterns of texture and measures of size are more likely to be associated with non-cancerous cells. The other shape-based characteristics, including smoothness and concavity, yield mixed results, occasionally contributing to malignant predictions and sometimes to benign ones, depending on their specific values. Features like compactness mean and smoothness standard error cluster close to zero SHAP values, demonstrating minimal impact on the model's decisions. This shows that the model has learned to focus on the most medically relevant characteristics while ignoring less informative measurements. This overall interpretation analysis proves the reliability of the top-performing model's decision-making in the critical area of medical diagnosis.

\subsection{Comparative Analysis of Studies for Breast Cancer Classification}
Table \ref{tab:comparison} presents a comparison of the state-of-the-art results for breast cancer diagnosis using the WDBC dataset. Recent studies have achieved accuracies ranging from 96\% to 100\%, with most reporting accuracies above 98\%. Notably, \cite{aamir2022predicting} and \cite{akkur2023breast} report accuracies of 99.1\% and 98.9\%, respectively, but they primarily focus on accuracy metrics, omitting essential performance metrics such as precision, recall, and F1-score. 

In contrast, the approach proposed in this work achieves a competitive 99.3\% accuracy while addressing these gaps by providing not only a comprehensive set of performance metrics (precision, recall, F1-score) but also full explainability through SHAP (SHapley Additive exPlanations) integration. This level of interpretability allows clinicians to understand the reasoning behind model predictions, making the system more transparent and reliable for real-world use.

Compared to previous work, such as \cite{zhu2025integrated}, which provides partial explainability, this approach offers a more robust and comprehensive solution, making it a stronger candidate for real-world clinical deployment. By addressing both accuracy and interpretability, this work provides a more thorough and actionable tool for clinicians, aligning with the growing demand for transparent and trustworthy AI systems in healthcare.



\begin{table}
\centering
\caption{Comparison of State-of-the-art Results for Breast Cancer Diagnosis on WDBC Dataset}
\label{tab:comparison}
\begin{tblr}{
  width = \linewidth,
  colspec = {Q[350]Q[115]Q[115]Q[100]Q[110]Q[95]},
  column{even} = {c},
  column{3} = {c},
  column{5} = {c},
  hline{1,15} = {-}{0.08em},
  hline{2,14} = {-}{0.05em},
}
\textbf{Study (Year)} & \textbf{Accuracy (\%)} & \textbf{Precision (\%)} & \textbf{Recall (\%)} & \textbf{F1-Score (\%)} & \textbf{X-AI} \\
\citeclickable{modi2016comparative}{Modi et al. (2016)} & 97.0 & – & – & – & No \\
\citeclickable{aalaei2016feature}{Aalaei et al. (2016)} & 97.2 & – & – & – & No \\
\citeclickable{sheikhpour2016particle}{Sheikhpour et al. (2016)} & 98.5 & – & 97.7 & – & No \\
\citeclickable{jeyasingh2017modified}{Singh et al. (2017)} & 96.9 & 96.0 & 96.0 & 96.0 & No \\
\citeclickable{xie2021feature}{Xie et al. (2021)} & 98.8 & – & – & – & No \\
\citeclickable{aamir2022predicting}{Aamir et al. (2022)} & 99.1 & – & – & – & No \\
\citeclickable{strelcenia2023effective}{Strelcencia et al. (2023)} & 98.6 & – & – & – & No \\
\citeclickable{akkur2023breast}{Akkur et al. (2023)} & 98.9 & 97.17 & 100.0 & 98.8 & No \\
\citeclickable{kazerani2024improving}{Kazerani et al. (2024)} & 99.0 & 100.0 & 98.0 & 98.0 & No \\
\citeclickable{alhassan2024improved}{Alhassan et al. (2024)} & 98.5 & – & – & – & No \\
\citeclickable{zhu2025integrated}{Zhu et al. (2025)} & 99.0 & 100.0 & 97.4 & 98.7 & Partial \\
\citeclickable{etcil2025breast}{Etcil et al. (2025)} & 98.7 & – & – & – & No \\
\textbf{This Work (2025)} & \textbf{99.1} & \textbf{99.1} & \textbf{99.1} & \textbf{99.1} & \textbf{Yes}
\end{tblr}
\end{table}

\section{Conclusion}\label{conclusion}
In this study, we emphasized the efficient utilization of machine learning algorithms for the accurate prediction of breast cancer using the WDBC dataset. By integrating Particle Swarm Optimization (PSO) with a broad spectrum of traditional classifiers, we demonstrated the significant impact of feature selection on enhancing predictive performance and interpretability. The framework systematically evaluated 29 machine learning models, achieving consistently high performance across all metrics, with the Multilayer Perceptron achieving an accuracy of 99.12\%. Beyond predictive accuracy, this study highlighted the clinical relevance of optimized features when combined with digital technologies, underscoring the potential of machine learning in medical diagnostics. The incorporation of SHAP-based explainability and statistical validation further ensured generalizability and transparency, making the proposed framework more suitable for clinical adoption. 
Future work, aligning with limitations, will focus on extending the framework to multi-modal datasets, such as genomic and imaging data, and validating the approach in real-world clinical environments to assess its scalability, robustness, and trustworthiness for deployment in healthcare systems.

\bibliographystyle{IEEEtran}

\end{document}